\documentclass[10pt,journal,compsoc]{IEEEtran}
\usepackage[pagebackref=true,breaklinks=true,letterpaper=true,colorlinks,citecolor=blue,linkcolor=blue,bookmarks=false]{hyperref}

\usepackage{xspace}
\usepackage{color}
\usepackage{multirow}
\usepackage{graphics} 
\usepackage{adjustbox}
\usepackage{amsmath}
\usepackage{amssymb}  
\usepackage{paralist} 
\usepackage{enumitem} 
\usepackage{booktabs} 

\graphicspath{{images/}}

\long\def\ignorethis#1{}

\definecolor{gray}{rgb}{0.35,0.35,0.35}
\definecolor{MyBlue}{rgb}{0,0.2,0.8}
\definecolor{MyRed}{rgb}{0.8,0.2,0}
\definecolor{MyGreen}{rgb}{0.0,0.5,0.1}
\definecolor{MyGray}{rgb}{0.4,0.4,0.4}

\def\red#1{\textcolor{red}{#1}}
\def\blue#1{\textcolor{blue}{#1}}

\def\first#1{\red{\textbf{#1}}}
\def\second#1{\blue{\underline{#1}}}

\newlength\paramargin
\newlength\figmargin
\newlength\secmargin

\usepackage{array}
\newcolumntype{L}[1]{>{\raggedright\let\newline\\\arraybackslash\hspace{0pt}}m{#1}}
\newcolumntype{C}[1]{>{\centering\let\newline\\\arraybackslash\hspace{0pt}}m{#1}}
\newcolumntype{R}[1]{>{\raggedleft\let\newline\\\arraybackslash\hspace{0pt}}m{#1}}

\def\eg{e.g.,~}

\def\etal{et~al.\xspace}

\newcommand{\secref}[1]{Section~\ref{sec:#1}}
\newcommand{\figref}[1]{Figure~\ref{fig:#1}}
\newcommand{\tabref}[1]{Table~\ref{tab:#1}}
\newcommand{\eqnref}[1]{\eqref{eq:#1}}

\ifCLASSOPTIONcompsoc
\usepackage[nocompress]{cite}
\else
\usepackage{cite}
\fi

\ifCLASSINFOpdf
\else
\fi
\begin{document}
\title{Fast and Accurate Image Super-Resolution with Deep Laplacian Pyramid Networks}

\author{Wei-Sheng Lai,
Jia-Bin Huang,
Narendra Ahuja,
and Ming-Hsuan Yang
\IEEEcompsocitemizethanks{
\IEEEcompsocthanksitem W.-S. Lai and M.-H. Yang are with the Department of Electrical and Engineering and Computer Science, University of California, Merced, CA, 95340. Email: $\{$wlai24$|$mhyang$\}$@ucmerced.edu
\IEEEcompsocthanksitem J.-B. Huang is with the Department of Electrical and Computer Engineering, Virginia Tech, VA 24060. Email: jbhuang@vt.edu
\IEEEcompsocthanksitem N. Ahuja is with the Department of Electrical and Computer Engineering, University of Illinois at Urbana-Champaign, IL 61801. Email: n-ahuja@illinois.edu
}
}

\markboth{}%
{Shell \MakeLowercase{\textit{et al.}}: Bare Demo of IEEEtran.cls for Computer Society Journals}
%



\IEEEtitleabstractindextext{%
\begin{abstract}
Convolutional neural networks have recently demonstrated high-quality reconstruction for single image super-resolution.
%
%
However, existing methods often require a large number of network parameters and entail heavy computational loads at runtime for generating high-accuracy super-resolution results.
%
%
In this paper, we propose the deep Laplacian Pyramid Super-Resolution Network for fast and accurate image super-resolution.
The proposed network progressively reconstructs the sub-band residuals of high-resolution images at multiple pyramid levels.
In contrast to existing methods that involve the bicubic interpolation for pre-processing (which results in large feature maps), the proposed method directly extracts features from the low-resolution input space and thereby entails low computational loads.
%
%
We train the proposed network with deep supervision using the robust Charbonnier loss functions and achieve high-quality image reconstruction.
%
%
Furthermore, we utilize the recursive layers to share parameters across as well as within pyramid levels, and thus drastically reduce the number of parameters.
%
%
Extensive quantitative and qualitative evaluations on benchmark datasets show that the proposed algorithm performs favorably against the state-of-the-art methods in terms of run-time and image quality.
\end{abstract}

}

\maketitle

\IEEEdisplaynontitleabstractindextext

%
\IEEEpeerreviewmaketitle


\IEEEraisesectionheading{\section{Introduction}}
\label{sec:introduction}
%
%
\IEEEPARstart{S}{ingle} image super-resolution (SR) aims to reconstruct a high-resolution (HR) image from one single low-resolution (LR) input image.
Example-based SR methods have demonstrated the state-of-the-art performance by learning a mapping from LR to HR image patches using large image datasets.
Numerous learning algorithms have been applied to learn such a mapping function, including dictionary learning~\cite{Yang-CVPR-2008,Yang-TIP-2010}, local linear regression~\cite{A+,Yang-ICCV-2013}, and random forest~\cite{RFL}, to name a few.

\begin{figure*}
	\centering
	\footnotesize
	\begin{tabular}{cc}
		\centering
		\begin{adjustbox}{valign=c}
			\begin{tabular}{c}
				\includegraphics[width=0.4\textwidth]{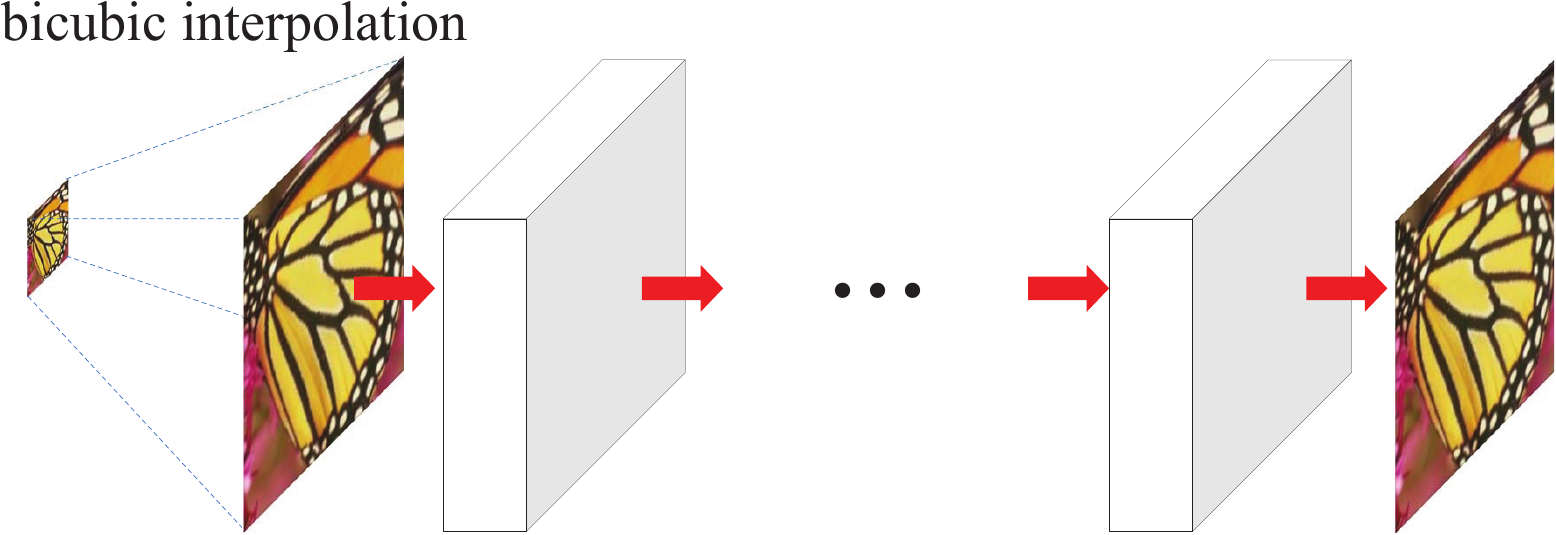}
				\\[1ex]
				(a) Pre-upsampling
				\\[7ex]
				\includegraphics[width=0.175\textwidth]{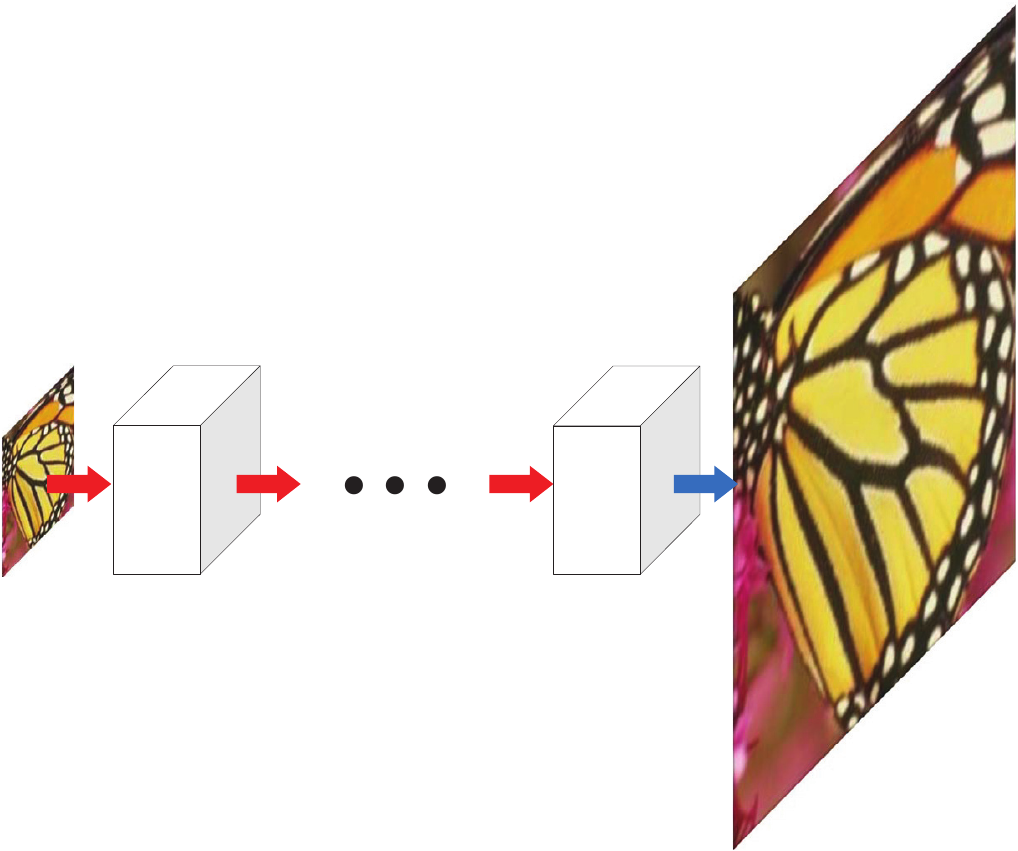}
				\\[1ex]
				(b) Post-upsampling
				\end{tabular}
		\end{adjustbox}
		& 
		\begin{adjustbox}{valign=c}
			\begin{tabular}{c}
				\includegraphics[width=0.5\textwidth]{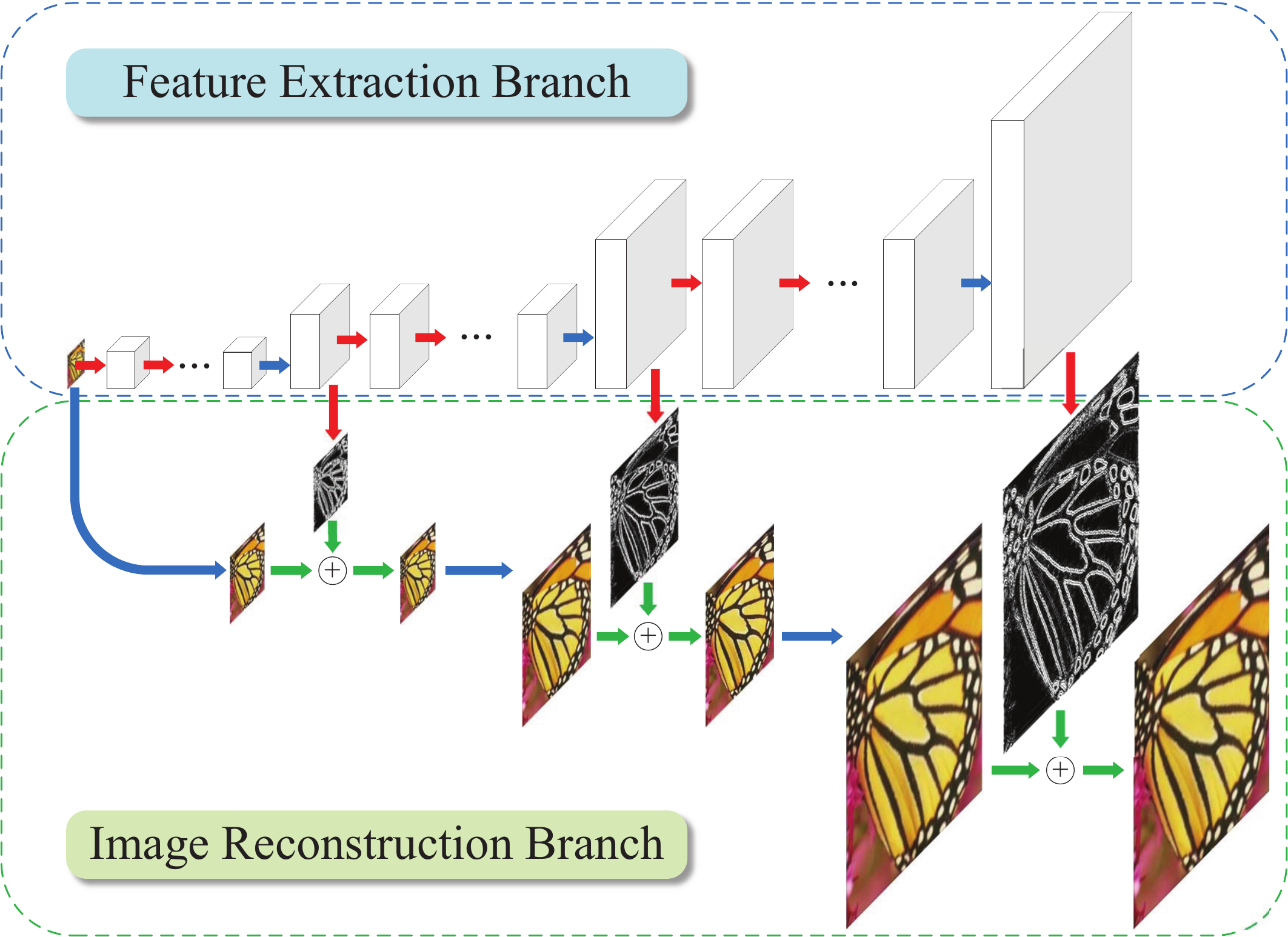}
				\\[1ex]
				(c) Progressive upsampling (ours)
			\end{tabular}
		\end{adjustbox}
	\end{tabular}
	\caption{
	\textbf{Comparisons of upsampling strategies in CNN-based SR algorithms.}
	Red arrows indicate convolutional layers. Blue arrows indicate transposed convolutions (upsampling), and green arrows denote element-wise addition operators.
	(a) Pre-upsampling based approaches (e.g., SRCNN~\cite{SRCNN}, VDSR~\cite{VDSR}, DRCN~\cite{DRCN}, DRRN~\cite{DRRN}) typically use the bicubic interpolation to upscale LR input images to the target spatial resolution before applying deep networks for prediction and reconstruction.
	(b) Post-upsampling based methods directly extract features from LR input images and use sub-pixel convolution~\cite{ESPCN} or transposed convolution~\cite{FSRCNN} for upsampling.
	(c) Progressive upsampling approach using the proposed Laplacian pyramid network reconstructs HR images in a coarse-to-fine manner.
	}
	\label{fig:upsampling}
\end{figure*}

%
Convolutional Neural Networks (CNNs) have been widely used in vision tasks ranging from object recognition~\cite{ResNet}, segmentation~\cite{FCN}, optical flow~\cite{flownet}, to super-resolution.
In~\cite{SRCNN}, Dong~\etal propose a Super-Resolution Convolutional Neural Network (SRCNN) to learn a nonlinear LR-to-HR mapping function. 
This network architecture has been extended to embed a sparse coding model~\cite{SCN}, increase network depth~\cite{VDSR}, or apply recursive layers~\cite{DRCN,DRRN}.
While these models are able to generate high-quality SR images, there remain three issues to be addressed. 
First, these methods use a pre-defined upsampling operator, \eg bicubic interpolation, to upscale an input LR image to the desired spatial resolution \emph{before} applying a network for predicting the details (\figref{upsampling}(a)).
This pre-upsampling step increases unnecessary computational cost and does not provide additional high-frequency information for reconstructing HR images.
Several algorithms accelerate the SRCNN by extracting features directly from the input LR images (\figref{upsampling}(b)) and replacing the pre-defined upsampling operator with sub-pixel convolution~\cite{ESPCN} or transposed convolution~\cite{FSRCNN} (also named as deconvolution in some literature).
These methods, however, use relatively small networks and cannot learn complicated mappings well due to the limited model capacity.
Second, existing methods optimize the networks with an $\mathcal{L}_2$ loss (i.e., mean squared error loss).
Since the same LR patch may have multiple corresponding HR patches and the $\mathcal{L}_2$ loss fails to capture the underlying multi-modal distributions of HR patches, the reconstructed HR images are often over-smoothed and inconsistent to human visual perception on natural images. 
Third, existing methods mainly reconstruct HR images in \textit{one upsampling step}, which makes learning mapping functions for large scaling factors (e.g., $8\times$) more difficult. 

To address these issues, we propose the deep Laplacian Pyramid Super-Resolution Network (LapSRN) to progressively reconstruct HR images in a coarse-to-fine fashion.
As shown in~\figref{upsampling}(c), our model consists of a feature extraction branch and an image reconstruction branch.
The feature extraction branch uses a cascade of convolutional layers to extract non-linear feature maps from LR input images.
We then apply a transposed convolutional layer for upsampling the feature maps to a finer level and use a convolutional layer to predict the sub-band residuals (i.e., the differences between the upsampled image and the ground truth HR image at the respective pyramid level).
The image reconstruction branch upsamples the LR images and takes the sub-band residuals from the feature extraction branch to efficiently reconstruct HR images through element-wise addition.
Our network architecture naturally accommodates deep supervision (i.e., supervisory signals can be applied simultaneously at each level of the pyramid) to guide the reconstruction of HR images.
Instead of using the $\mathcal{L}_2$ loss function, we propose to train the network with the robust Charbonnier loss functions to better handle outliers and improve the performance.
While both feature extraction and image reconstruction branches have multiple levels, we train the network in an end-to-end fashion without stage-wise optimization.

Our algorithm differs from existing CNN-based methods in the following three aspects: 
\begin{enumerate}
\item \textbf{Accuracy}. 
Instead of using a pre-defined upsampling operation, our network jointly optimizes the deep convolutional layers and upsampling filters for both images and feature maps by minimizing the Charbonnier loss function.
As a result, our model has a large capacity to learn complicated mappings and effectively reduces the undesired artifacts caused by spatial aliasing. 
\item \textbf{Speed}.
Our LapSRN accommodates both fast processing speed and high capacity of deep networks.
Experimental results demonstrate that our method is faster than several CNN-based super-resolution models, e.g., VDSR~\cite{VDSR}, DRCN~\cite{DRCN}, and DRRN~\cite{DRRN}.
The proposed model achieves real-time performance as FSRCNN~\cite{FSRCNN} while generating significantly better reconstruction accuracy.
\item \textbf{Progressive reconstruction}. 
Our model generates multiple intermediate SR predictions in \emph{one} feed-forward pass through progressive reconstruction.
This characteristic renders our method applicable to a wide range of tasks that require resource-aware adaptability. 
For example, the same network can be used to enhance the spatial resolution of videos depending on the available computational resources.
For scenarios with limited computing resources, our $8\times$ model can still perform 2$\times$ or $4\times$ SR by simply bypassing the computation of residuals at finer levels. 
Existing CNN-based methods, however, do not offer such flexibility.

\end{enumerate}

In this work, we make the following extensions to improve our early results~\cite{LapSRN} substantially:
\begin{enumerate}
\item \textbf{Parameter sharing}.
We re-design our network architecture to share parameters \textit{across} pyramid levels and \textit{within} the feature extraction sub-network via recursion.
Through parameter sharing, we reduce $73\%$ of the network parameters while achieving better reconstruction accuracy on benchmark datasets. 
\item \textbf{Local skip connections}.
We systematically analyze three different approaches for applying local skip connections in the proposed model.
By leveraging proper skip connections to alleviate the gradient vanishing and explosion problems, we are able to train an 84-layer network to achieve the state-of-the-art performance.

\item \textbf{Multi-scale training}. 
Unlike in the preliminary work where we train three different models for handling $2\times$, $4\times$ and $8\times$ SR, respectively, we train one \textit{single} model to handle \textit{multiple} upsampling scales.
The multi-scale model learns the inter-scale correlation and improves the reconstruction accuracy against single-scale models.
We refer to our multi-scale model as MS-LapSRN.
\end{enumerate}

\begin{table*}
	\centering
	\caption{
		\textbf{Feature-by-feature comparisons of CNN-based SR algorithms.}
		Methods with direct reconstruction performs one-step upsampling from the LR to HR space, while progressive reconstruction predicts HR images in multiple upsampling steps.
		Depth represents the number of convolutional and transposed convolutional layers in the longest path from input to output for $4\times$ SR.
		Global residual learning (GRL) indicates that the network learns the difference between the ground truth HR image and the upsampled (i.e., using bicubic interpolation or learned filters) LR images.
		Local residual learning (LRL) stands for the local skip connections between intermediate convolutional layers.
	}
	\vspace{-0.1cm}
	\begin{tabular}{rccccccccc}
	\toprule
	\multirow{2}{*}{Method} & 
	\multirow{2}{*}{Input} & 
	\multirow{2}{*}{Reconstruction} &
	\multirow{2}{*}{Depth} & 
	\multirow{2}{*}{Filters} & 
	\multirow{2}{*}{Parameters} & 
	\multirow{2}{*}{GRL} &
	\multirow{2}{*}{LRL} &
	Multi-scale &
	\multirow{2}{*}{Loss function}
	\\
	&
	&
	&
	&
	&
	&
	&
	&
	training &
	\\
	\midrule
	SRCNN~\cite{SRCNN} 
	& LR + bicubic & Direct
	& 3 & 64 & 57k & & & & $\mathcal{L}_2$ \\
	FSRCNN~\cite{FSRCNN} 
	& LR & Direct 
	& 8 & 56 & 12k & & & & $\mathcal{L}_2$ \\
	ESPCN~\cite{ESPCN} 
	& LR & Direct 
	& 3 & 64 & 20k & & & & $\mathcal{L}_2$ \\
	SCN~\cite{SCN} 
	& LR + bicubic & Progressive 
	& 10 & 128 & 42k & & & & $\mathcal{L}_2$ \\
	VDSR~\cite{VDSR} 
	& LR + bicubic & Direct 
	& 20 & 64 & 665k & \checkmark & & \checkmark & $\mathcal{L}_2$ \\
	DRCN~\cite{DRCN} 
	& LR + bicubic & Direct 
	& 20 & 256 & 1775k & \checkmark & & & $\mathcal{L}_2$ \\
	DRRN~\cite{DRRN} 
	& LR + bicubic & Direct 
	& 52 & 128 & 297k & \checkmark & \checkmark & \checkmark & $\mathcal{L}_2$ \\
	%
	MDSR~\cite{MDSR}
	& LR & Direct
	& 162 & 64 & 8000k & & \checkmark & \checkmark & Charbonnier \\
	LapSRN~\cite{LapSRN}
	& LR & Progressive 
	& 24 & 64 & 812k & \checkmark & & & Charbonnier \\
	MS-LapSRN (ours)
	& LR & Progressive
	& 84 & 64 & 222k & \checkmark & \checkmark & \checkmark & Charbonnier \\
	\bottomrule
	\end{tabular}
	\label{tab:CNN-compare}
\end{table*}


\section{Related Work}

Single-image super-resolution has been extensively studied in the literature.
Here we focus our discussion on recent example-based and CNN-based approaches.

\subsection{SR based on internal databases}
Several methods~\cite{Freedman-TOG-2011,Yang-ACCV-2010,Glasner-ICCV-2009} exploit the self-similarity property in natural images and construct LR-HR patch pairs based on the scale-space pyramid of the LR input image.
While internal databases contain more relevant training patches than external image datasets, the number of LR-HR patch pairs may not be sufficient to cover large textural appearance variations in an image.
Singh~\etal~\cite{Singh-ACCV-2014} decompose patches into directional frequency sub-bands and determine better matches in each sub-band pyramid independently.
In~\cite{SelfExSR}, Huang~\etal extend the patch search space to accommodate the affine and perspective deformation.
The SR methods based on internal databases are typically slow due to the heavy computational cost of patch searches in the scale-space pyramid.
Such drawbacks make these approaches less feasible for applications that require computational efficiency.

\subsection{SR based on external databases}
Numerous SR methods learn the LR-HR mapping with image pairs collected from external databases using supervised learning algorithms, such as nearest neighbor~\cite{Freeman-CGA-2002}, manifold embedding~\cite{Bevilacqua-BMVC-2012,Chang-CVPR-2004}, kernel ridge regression~\cite{Kim-PAMI-2010}, and sparse representation~\cite{Yang-CVPR-2008,Yang-TIP-2010,Zeyde-2010}.
Instead of directly modeling the complex patch space over the entire database, recent methods partition the image set by K-means~\cite{Yang-ICCV-2013}, sparse dictionary~\cite{ANR,A+} or random forest~\cite{RFL}, and learn locally linear regressors for each cluster.
While these approaches are effective and efficient, the extracted features and mapping functions are hand-designed, which may not be optimal for generating high-quality SR images.

\subsection{Convolutional neural networks based SR} 
CNN-based SR methods have demonstrated state-of-the-art results by jointly optimizing the feature extraction, non-linear mapping, and image reconstruction stages in an end-to-end manner.
The VDSR network~\cite{VDSR} shows significant improvement over the SRCNN method~\cite{SRCNN} by increasing the network depth from 3 to 20 convolutional layers.
To facilitate training a deeper model with a fast convergence speed, the VDSR method adopts the global residual learning paradigm to predict the differences between the ground truth HR image and the bicubic upsampled LR image instead of the actual pixel values.
Wang~\etal~\cite{SCN} combine the domain knowledge of sparse coding with a deep CNN and train a cascade network (SCN) to upsample images progressively.
In~\cite{DRCN}, Kim~\etal propose a network with multiple recursive layers (DRCN) with up to 16 recursions.
The DRRN approach~\cite{DRRN} further trains a 52-layer network by extending the local residual learning approach of the ResNet~\cite{ResNet} with deep recursion.
We note that the above methods use bicubic interpolation to pre-upsample input LR images \emph{before} feeding into the deep networks, which increases the computational cost and requires a large amount of memory.

To achieve real-time speed, the ESPCN method~\cite{ESPCN} extracts feature maps in the LR space and replaces the bicubic upsampling operation with an efficient sub-pixel convolution (i.e., pixel shuffling).
The FSRCNN method~\cite{FSRCNN} adopts a similar idea and uses a hourglass-shaped CNN with transposed convolutional layers for upsampling. 
As a trade-off of speed, both ESPCN~\cite{ESPCN} and FSRCNN~\cite{FSRCNN} have limited network capacities for learning complex mappings.
Furthermore, these methods upsample images or features in \emph{one upsampling step} and use only one supervisory signal from the target upsampling scale.
Such a design often causes difficulties in training models for large upsampling scales (e.g., $4\times$ or $8\times$).
%
%
In contrast, our model progressively upsamples input images on \emph{multiple} pyramid levels and use \emph{multiple} losses to guide the prediction of sub-band residuals at each level, which leads to accurate reconstruction, particularly for large upsampling scales.

All the above CNN-based SR methods optimize networks with the $\mathcal{L}_2$ loss function, which often leads to over-smooth results that do not correlate well with human perception.
We demonstrate that the proposed deep network with the robust Charbonnier loss function better handles outliers and improves the SR performance over the $\mathcal{L}_2$ loss function.
Most recently, Lim~\etal~\cite{MDSR} propose a multi-scale deep SR model (MDSR) by extending ESPCN~\cite{ESPCN} with three branches for scale-specific upsampling but sharing most of the parameters across different scales.
The MDSR method is trained on a high-resolution
DIV2K~\cite{NTIRE} dataset (800 training images of 2k resolution), and achieves the state-of-the-art performance.
\tabref{CNN-compare} shows the main components of the existing CNN-based SR methods.
%
The preliminary method of this work, i.e., LapSRN~\cite{LapSRN}, and the proposed MS-LapSRN algorithm are listed in the last two rows.

\subsection{Laplacian pyramid} 
\label{sec:related_lapalcian}
The Laplacian pyramid has been widely used in several vision tasks, including image blending~\cite{Burt-1983}, texture synthesis~\cite{Heeger-TOG-1995}, edge-aware filtering~\cite{Paris-TOG-2011} and semantic segmentation~\cite{Ghiasi-ECCV-2016}.
Denton~\etal~\cite{LAPGAN} propose a generative adversarial network based on a Laplacian pyramid framework (LAPGAN) to generate realistic images, which is the most related to our work.
However, the proposed LapSRN differs from LAPGAN in two aspects.

First, the \textit{objectives} of the two models are different.
The LAPGAN is a generative model which is designed to synthesize diverse natural images from random noise and sample inputs.
On the contrary, the proposed LapSRN is a super-resolution model that predicts a particular HR image based on the given LR image and upsampling scale factor.
The LAPGAN uses a cross-entropy loss function to encourage the output images to respect the data distribution of the training datasets. 
In contrast, we use the Charbonnier penalty function to penalize the deviation of the SR prediction from the ground truth HR images.

\begin{figure}
	\centering
	\includegraphics[width=0.7\linewidth]{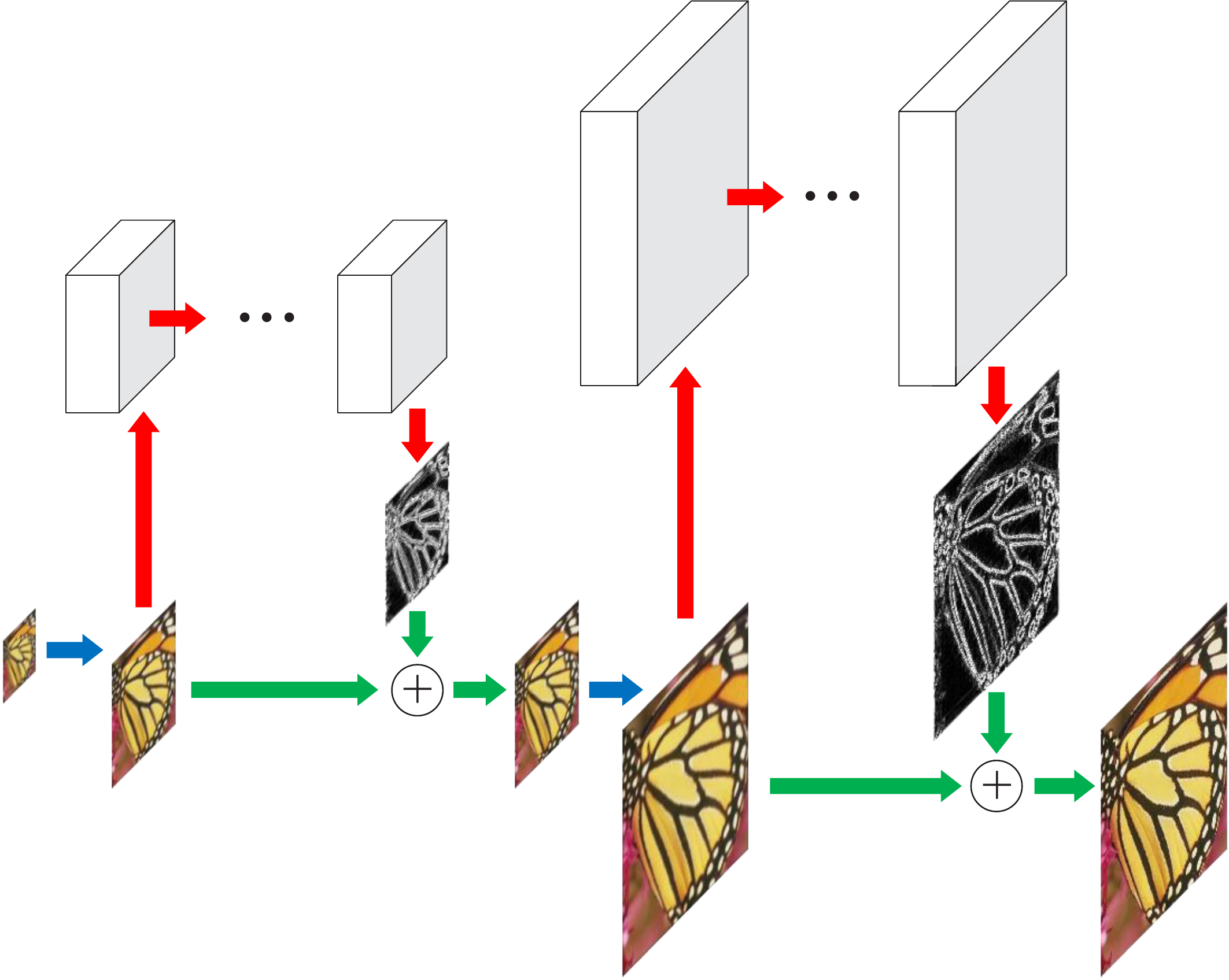}
	\caption{
	\textbf{Generative network of LAPGAN~\cite{LAPGAN}.}
	The LAPGAN first upsamples the input images before applying convolution for predicting residuals at each pyramid level.
	}
	\label{fig:LAPGAN}
\end{figure}

Second, the differences in \textit{architecture designs} result in disparate inference speed and network capacities.
As shown in~\figref{LAPGAN}, the LAPGAN upsamples input images \textit{before} applying convolution at each level, while our LapSRN extracts features directly from the LR space and upscales images at the end of each level.
Our network design effectively alleviates the computational cost and increases the size of receptive fields.
In addition, the convolutional layers at each level in our LapSRN are \emph{connected} through multi-channel transposed convolutional layers.
The residual images at a higher level are therefore predicted by a deeper network with shared feature representations at lower levels.
The shared features at lower levels increase the non-linearity at finer convolutional layers to learn complex mappings.

\subsection{Adversarial training} 
The Generative Adversarial Networks (GANs)~\cite{GAN} have been applied to several image reconstruction and synthesis problems, including image inpainting~\cite{Pathak-CVPR-2016}, face completion~\cite{Li-CVPR-2017}, and face super-resolution~\cite{Yu-ECCV-2016}.
Ledig~\etal~\cite{SRGAN} adopt the GAN framework for learning natural image super-resolution.
The ResNet~\cite{ResNet} architecture is used as the generative network and train the network using the combination of the $\mathcal{L}_2$ loss, perceptual loss~\cite{Johnson-ECCV-2016}, and adversarial loss.
The SR results may have lower PSNR but are visually plausible.
Note that our LapSRN can be easily extended to incorporate adversarial training.
We present experimental results on training with the adversarial loss in~\secref{adversarial}.


\section{Deep Laplacian Pyramid Network for SR}
\label{sec:framework}
%
In this section, we describe the design methodology of the proposed LapSRN, including the network architecture, parameter sharing, loss functions, multi-scale training strategy, and details of implementation as well as network training.

\begin{figure*}
	\centering
	\includegraphics[width=0.9\linewidth]{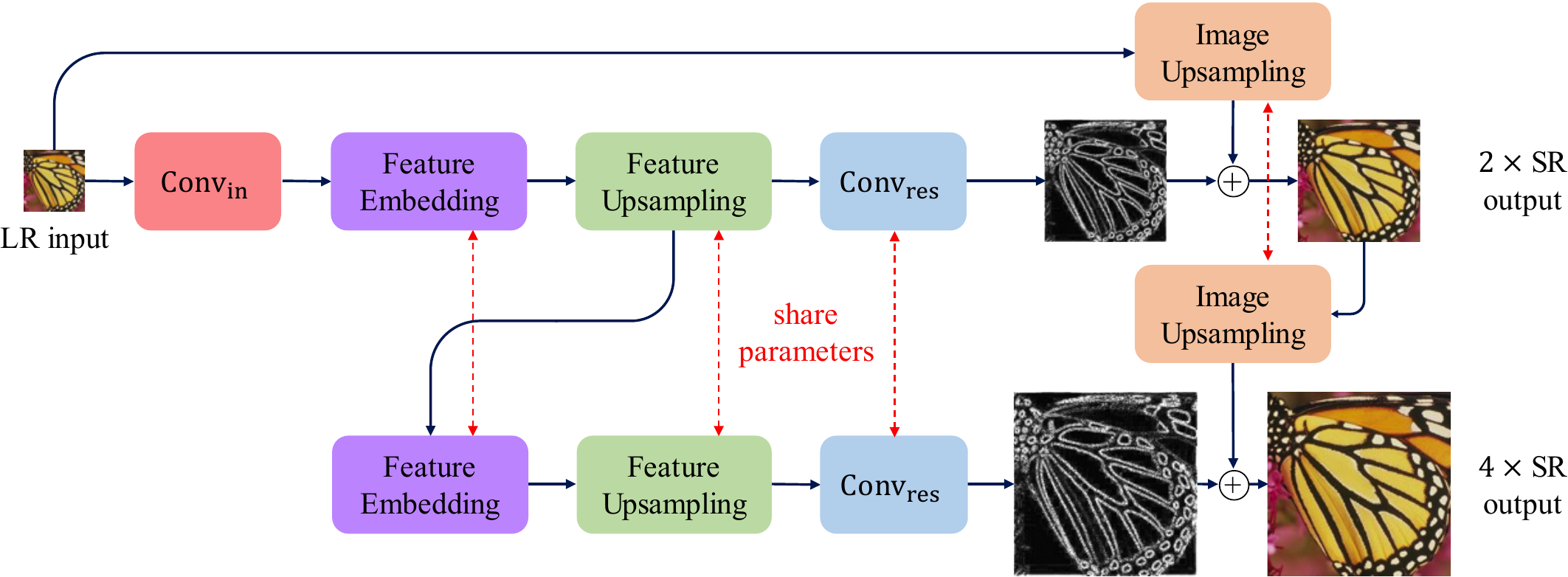} 
	\caption{
		\textbf{Detailed network architecture of the proposed LapSRN.}
		At each pyramid level, our model consists of a feature embedding sub-network for extracting non-linear features, transposed convolutional layers for upsampling feature maps and images, and a convolutional layer for predicting the sub-band residuals.
		As the network structure at each level is highly similar, we share the weights of those components across pyramid levels to reduce the number of network parameters.
	}
	\label{fig:parameter_sharing}
\end{figure*}

\subsection{Network architecture}
\label{sec:architecture}
%
We construct our network based on the Laplacian pyramid framework.
Our model takes an LR image as input (rather than an upscaled version of the LR image) and progressively predicts residual images on the $\log_2 S$ pyramid levels, where $S$ is the upsampling scale factor.
For example, our network consists of $3$ pyramid levels for super-resolving an LR image at a scale factor of $8$.
Our model consists of two branches: (1) feature extraction and (2) image reconstruction.

\subsubsection{Feature extraction branch} 
As illustrated in~\figref{upsampling}(c) and~\figref{parameter_sharing}, the feature extraction branch consists of 
(1) a feature embedding sub-network for transforming high-dimensional non-linear feature maps, 
(2) a transposed convolutional layer for upsampling the extracted features by a scale of 2, and 
(3) a convolutional layer ($\text{Conv}_{\text{res}}$) for predicting the sub-band residual image.
The first pyramid level has an additional convolutional layer  ($\text{Conv}_{\text{in}}$) to extract high-dimensional feature maps from the input LR image.
At other levels, the feature embedding sub-network directly transforms features from the upscaled feature maps at the previous pyramid level.
Unlike the design of the LAPGAN in~\figref{LAPGAN}, we do not \emph{collapse} the feature maps into an image before feeding into the next level.
Therefore, the feature representations at lower levels are connected to higher levels and thus can increase the non-linearity of the network to learn complex mappings at the finer levels.
Note that we perform the feature extraction at the \emph{coarse} resolution and generate feature maps at the \emph{finer} resolution with only one transposed convolutional layer.
In contrast to existing networks (e.g., \cite{VDSR,DRRN}) that perform all feature extraction and reconstruction at the finest resolution, our network design significantly reduces the computational complexity.

\subsubsection{Image reconstruction branch} 
At level $s$, the input image is upsampled by a scale of 2 with a transposed convolutional layer, which is initialized with a $4 \times 4$ bilinear kernel.
We then combine the upsampled image (using element-wise summation) with the predicted residual image to generate a high-resolution output image.
The reconstructed HR image at level $s$ is then used as an input for the image reconstruction branch at level $s+1$.
The entire network is a cascade of CNNs with the same structure at each level.
We jointly optimize the upsampling layer with all other layers to learn better a upsampling function. 

\subsection{Feature embedding sub-network}
In our preliminary work~\cite{LapSRN}, we use a stack of multiple convolutional layers as our feature embedding sub-network.
In addition, we learn distinct sets of convolutional filters for feature transforming and upsampling at different pyramid levels.
Consequently, the number of network parameters increases with the depth of the feature embedding sub-network and the upsampling scales, e.g., the $4\times$ SR model has about twice number of parameters than the $2\times$ SR model.
In this work, we explore two directions to reduce the network parameters of LapSRN.

\subsubsection{Parameter sharing across pyramid levels}
Our first strategy is to share the network parameters \emph{across} pyramid levels as the network at each level shares the same structure and the task (i.e., predicting the residual images at $2\times$ resolution).
As shown in~\figref{parameter_sharing}, we share the parameters of the feature embedding sub-network, upsampling layers, and the residual prediction layers across all the pyramid levels.
As a result, the number of network parameters is independent of the upsampling scales.
%
We can use a single set of parameters to construct multi-level LapSRN models to handle different upsampling scales.

\subsubsection{Parameter sharing within pyramid level}
\label{sec:parameter_sharing}
Our second strategy is to share the network parameters \emph{within} each pyramid level.
Specifically, we extend the feature embedding sub-network using deeply recursive layers to effectively increase the network depth without increasing the number of parameters.
The design of recursive layers has been adopted by several recent CNN-based SR approaches.
The DRCN method~\cite{DRCN} applies a \emph{single} convolutional layer repeatedly up to 16 times.
However, with a large number of filters (i.e., 256 filters), the DRCN is memory-demanding and slow at runtime.
Instead of reusing the weights of a single convolutional layer, the DRRN~\cite{DRRN} method shares the weights of a \emph{block} (2 convolutional layers with 128 filters).
In addition, the DRRN introduces a variant of local residual learning from the ResNet~\cite{ResNet}.
Specifically, the identity branch of the ResNet comes from the output of the \emph{previous} block, while the identity branch of the DRRN comes from the input of the \emph{first} block.
Such a local skip connection in the DRRN creates multiple short paths from input to output and thereby effectively alleviates the gradient vanishing and exploding problems.
Therefore, DRRN has 52 convolutional layers with only 297k parameters.

\begin{figure}
	\centering
	\begin{tabular}{C{2.3cm}C{2.8cm}C{2.6cm}}
		\includegraphics[height=0.3\textwidth]{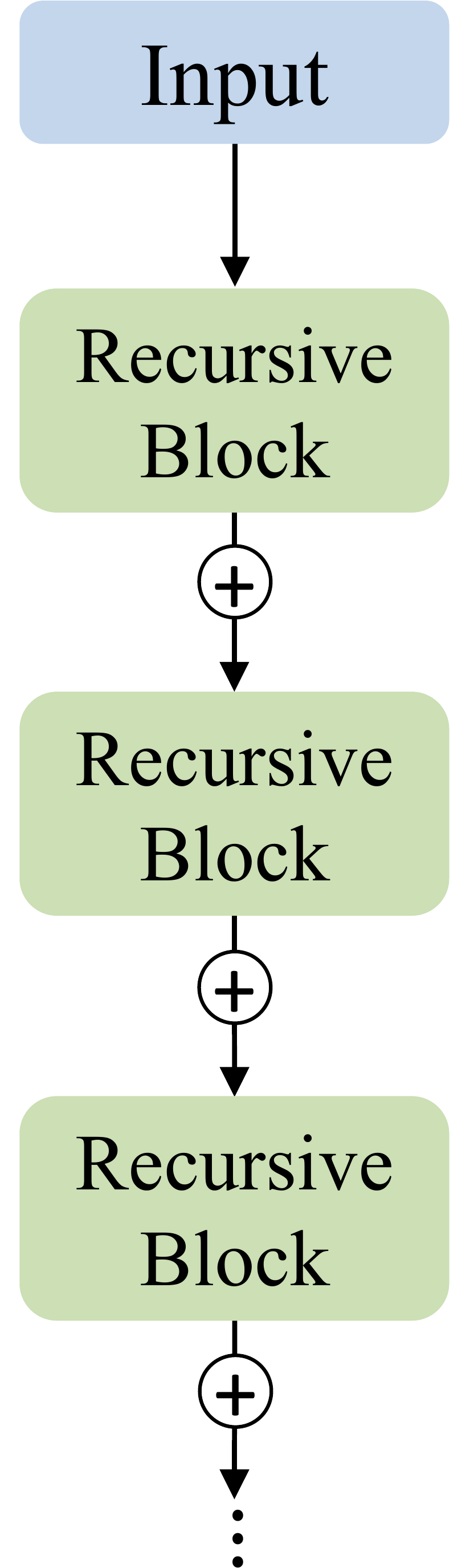}
		&
		\includegraphics[height=0.3\textwidth]{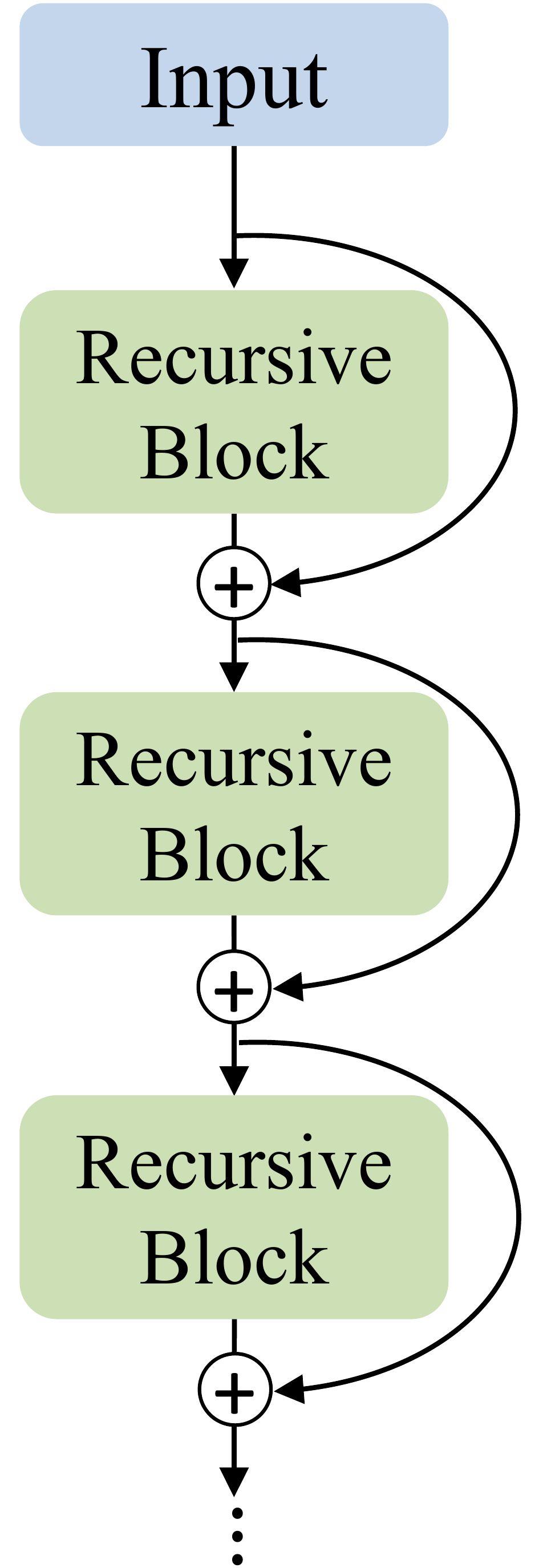}
		&
		\includegraphics[height=0.3\textwidth]{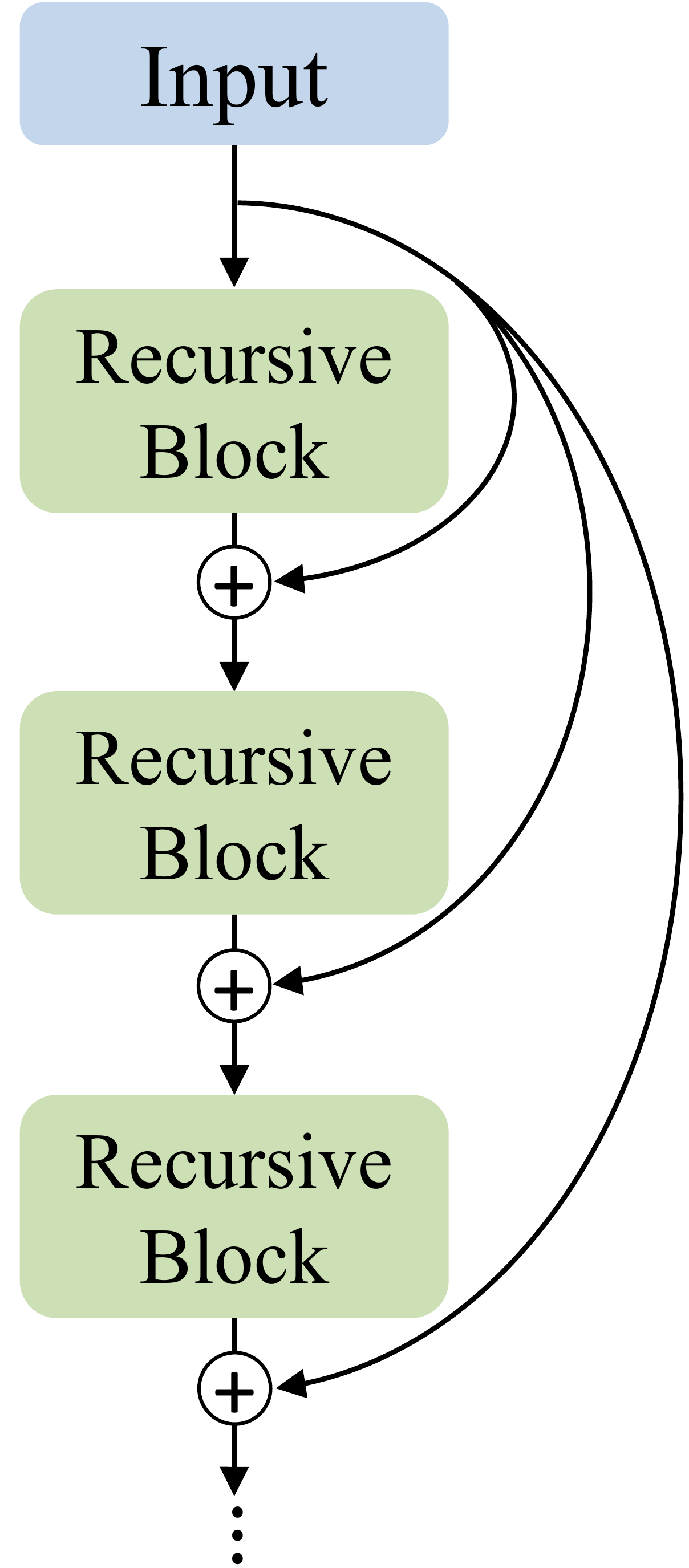} 
		\\
		(a) No skip connection &
		(b) Distinct-source skip connection & 
		(c) Shared-source skip connection
	\end{tabular}
	\caption{
		\textbf{Local residual learning.}
		We explore three different ways of local skip connection in the feature embedding sub-network of the LapSRN for training deeper models. 
	}
	\label{fig:LRL}
\end{figure}

\begin{figure}
	\centering
	\hspace{1.8cm}
	\includegraphics[width=0.6\linewidth]{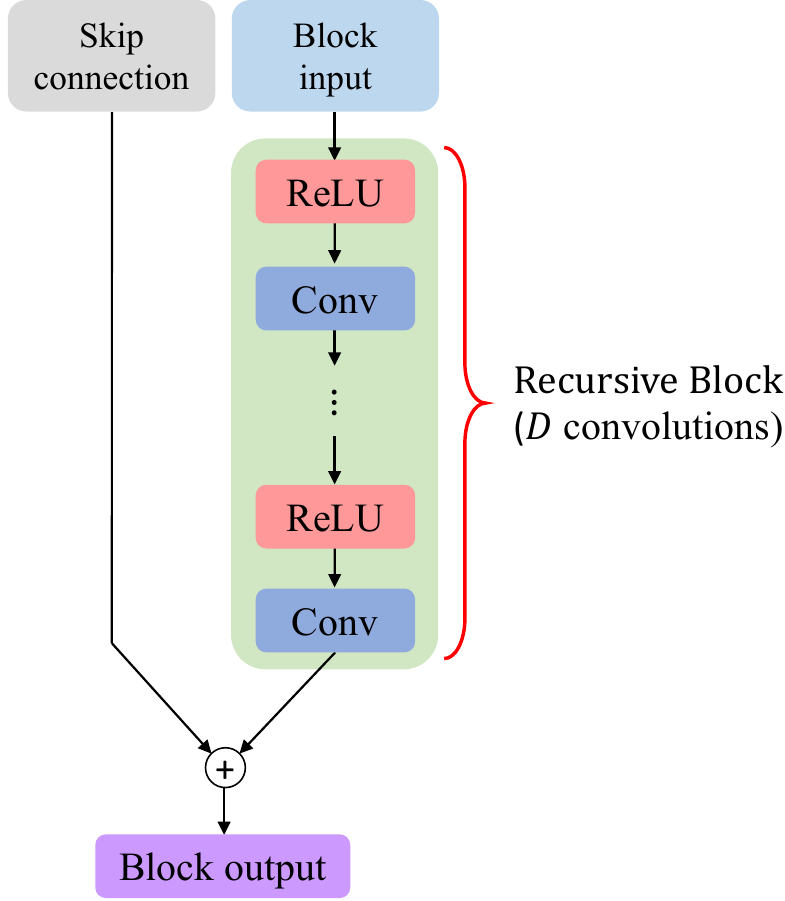} 
	\caption{
		\textbf{Structure of our recursive block.}
		There are $D$ convolutional layers in a recursive block.
		The weights of convolutional layers are distinct within the block but shared among all recursive blocks.
		We use the pre-activation structure~\cite{He-ECCV-2016} without the batch normalization layer.
	}
	\label{fig:recursive_block}
\end{figure}

In the proposed LapSRN, the feature embedding sub-network has $R$ recursive blocks.
Each recursive block has $D$ distinct convolutional layers, which controls the number of parameters in the entire model.
The weights of the $D$ convolutional layers are shared among the recursive blocks.
Given an upsampling scale factor $S$, the depth of the LapSRN can be computed by:
\begin{equation}
	\text{depth} = (D \times R + 1) \times L + 2, 
	\label{eq:depth}
\end{equation}
where $L = \log_2 S$. 
The $1$ within the parentheses represents the transposed convolutional layers, and the $2$ at the end of~\eqnref{depth} represents the first convolutional layer applied on input images and the last convolutional layer for predicting residuals.
Here we define the depth of a network as the longest path from input to output.

\subsubsection{Local residual learning}
\label{sec:LRL}
As the gradient vanishing and exploding problem are common issues when training deep models, we explore three different methods of local residual learning in our feature embedding sub-network to stabilize our training process:
\begin{enumerate}
\item \textbf{No skip connection}: 
A plain network without any local skip connection.
We denote our LapSRN without skip connections as $\text{LapSRN}_{\textbf{NS}}$.
\item \textbf{Distinct-source skip connection}: 
The ResNet-style local skip connection. 
We denote our LapSRN with such skip connections as $\text{LapSRN}_{\textbf{DS}}$.
\item \textbf{Shared-source skip connection}: 
The local skip connection introduced by DRRN~\cite{DRRN}. 
We denote our LapSRN with such skip connections as $\text{LapSRN}_{\textbf{SS}}$.
\end{enumerate}
We illustrate the three local residual learning methods in~\figref{LRL} and the detailed structure of our recursive block in~\figref{recursive_block}.
We use the pre-activation structure~\cite{He-ECCV-2016} without the batch normalization layer in our recursive block.

\subsection{Loss function}
\label{sec:loss}
Let $x$ be the input LR image and $\theta$ be the set of network parameters to be optimized.
Our goal is to learn a mapping function $f$ for generating an HR image $\hat{y} = f(x; \theta)$ that is as similar to the ground truth HR image $y$ as possible.
We denote the residual image at level $l$ by $\hat{r}_l$, the upscaled LR image by $x_l$ and the corresponding HR images by $\hat{y}_l$.
The desired output HR images at level $l$ is modeled by $\hat{y}_l = x_l + \hat{r}_l$.
We use the bicubic downsampling to resize the ground truth HR image $y$ to $y_l$ at each level.
Instead of minimizing the mean square errors between $\hat{y}_l$ and $y_l$, we propose to use a robust loss function to handle outliers.
The overall loss function is defined as:
\begin{align}
\mathcal{L}_S(y, \hat{y}; \theta) 
=&~ \frac{1}{N} \sum_{i=1}^N\sum_{l=1}^L \rho\left( y_l^{(i)} - \hat{y}_l^{(i)} \right) \nonumber\\
=&~ \frac{1}{N} \sum_{i=1}^N\sum_{l=1}^L \rho\left( (y_l^{(i)} - x_l^{(i)}) - \hat{r}_l^{(i)} \right),
\end{align}
where $\rho(x) = \sqrt{x^2 + \epsilon^2}$ is the Charbonnier penalty function (a differentiable variant of $\mathcal{L}_1$ norm)~\cite{Bruhn-IJCV-2005}, 
$N$ is the number of training samples in each batch,
$S$ is the target upsampling scale factor, 
and $L = \log_2S$ is the number of pyramid levels in our model.
We empirically set $\epsilon$ to $1e-3$.

In the proposed LapSRN, each level $s$ has its own loss function and the corresponding ground truth HR image $y_s$.
This multi-loss structure resembles the deeply-supervised networks for classification~\cite{Lee-AISTATS-2015} and edge detection~\cite{Xie-CVPR-2015}.
%
%
%
{The deep multi-scale supervision guides the network to reconstruct HR images in a coarse-to-fine fashion and reduce spatial aliasing artifacts.}

\subsection{Multi-scale training}
\label{sec:multiscale}
The multi-scales SR models (i.e., trained with samples from multiple upsampling scales simultaneously) have been shown more effective than single-scale models as SR tasks have inter-scale correlations.
For pre-upsampling based SR methods (e.g., VDSR~\cite{VDSR} and DRRN~\cite{DRRN}), the input and output of the network have the same spatial resolution, and the outputs of different upsampling scales are generated from the \emph{same} layer of the network.
In the proposed LapSRN, samples of different upsampling scales are generated from \emph{different} layers and have different spatial resolutions.
In this work, we use $2\times$, $4\times$, and $8\times$ SR samples to train a multi-scale LapSRN model.
We construct a 3-level LapSRN model and minimize the combination of loss functions from three different scales:
\begin{align}
\mathcal{L}(y, \hat{y}; \theta) = \sum_{S \in \{2, 4, 8\}} \mathcal{L}_S(y, \hat{y}; \theta).
\end{align}
We note that the pre-upsampling based SR methods could apply scale augmentation for arbitrary upsampling scales, while in our LapSRN, the upsampling scales for training are limited to $2^n\times$ SR where $n$ is an integer.

\subsection{Implementation and training details}
%
{In the proposed LapSRN, we use 64 filters in all convolutional layers
except the first layer applied on the input LR image, the layers for predicting residuals, and the image upsampling layer.}
{The filter size of the convolutional and transposed convolutional layers are $3 \times 3$ and $4 \times 4$, respectively.}
We pad zeros around the boundaries before applying convolution to keep the size of all feature maps the same as the input of each level.
We initialize the convolutional filters using the method of He~\etal~\cite{He-ICCV-2015} and use the leaky rectified linear units (LReLUs)~\cite{Maas-ICML-2013} with a negative slope of 0.2 as the non-linear activation function.
We use 91 images from Yang~\etal~\cite{Yang-TIP-2010} and 200 images from the training set of the Berkeley Segmentation Dataset~\cite{BSDS} as our training data.
The training dataset of 291 images is commonly used in the state-of-the-art SR methods~\cite{VDSR,RFL,LapSRN,DRRN}.
We use a batch size of 64 and crop the size of HR patches to $128 \times 128$.
An epoch has $1,000$ iterations of back-propagation. 
We augment the training data in three ways: 
(1) \emph{Scaling}: randomly downscale images between $[0.5, 1.0]$;
(2) \emph{Rotation}: randomly rotate image by $90^\circ$, $180^\circ$, or $270^\circ$;
(3) \emph{Flipping}: flip images horizontally with a probability of $0.5$. 
Following the training protocol of existing methods~\cite{SRCNN,VDSR,DRRN}, we generate the LR training patches using the bicubic downsampling.
We use the MatConvNet toolbox~\cite{Vedaldi-ACMMM-2015} and train our model using the Stochastic Gradient Descent (SGD) solver.
In addition, we set the momentum to $0.9$ and the weight decay to $1e-4$. 
The learning rate is initialized to $1e-5$ for all layers and decreased by a factor of 2 for every 100 epochs.
%


\section{Discussions and Analysis}
\label{sec:analysis}
In this section, we first validate the contributions of different components in the proposed network.
We then discuss the effect of local residual learning and parameter sharing in our feature embedding sub-network.
Finally, we analyze the performance of multi-scale training strategy.

\subsection{Model design} 
We train a LapSRN model with 5 convolutional layers (without parameters sharing and the recursive layers) at each pyramid level to analyze the performance of pyramid network structure, global residual learning, robust loss functions, and multi-scale supervision.

\subsubsection{Pyramid structure}
\begin{figure}
	\scriptsize
	\centering
	\begin{tabular}{cc}
		\hspace{-0.45cm}
		\begin{adjustbox}{valign=t}
			\begin{tabular}{c}
				\includegraphics[width=0.29\columnwidth]{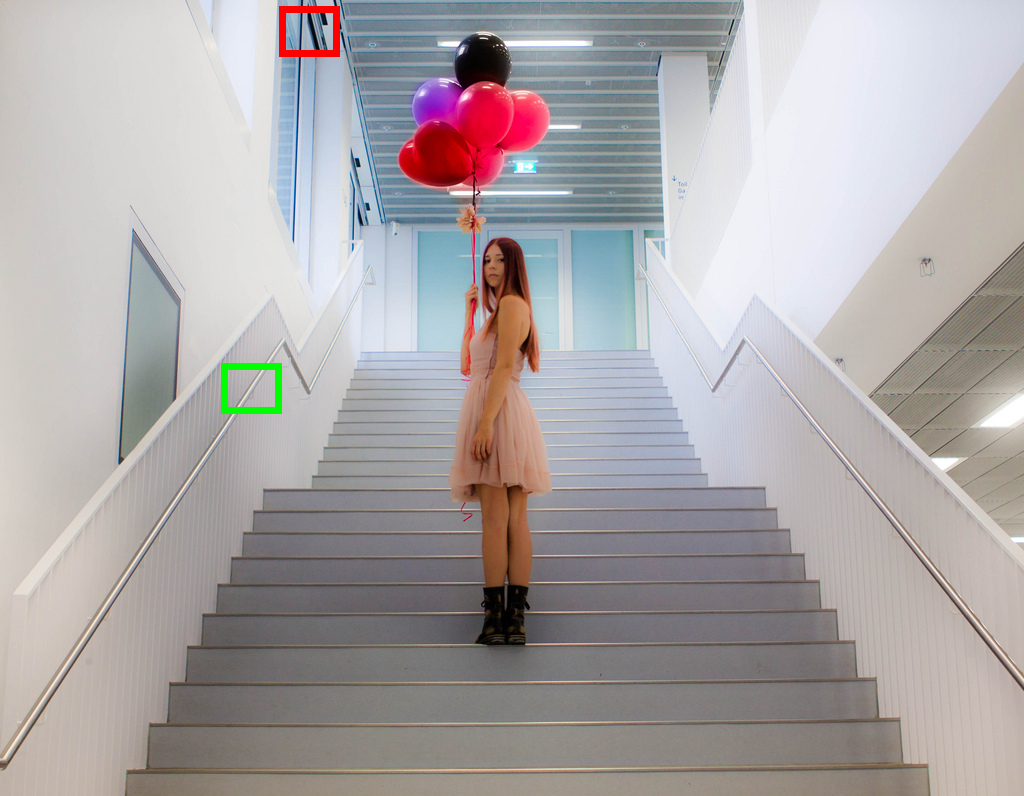}
				\\
				(a)
			\end{tabular}
		\end{adjustbox}
		\hspace{-0.45cm}
		\begin{adjustbox}{valign=t}
			\begin{tabular}{ccccc}
				\includegraphics[width=0.13\columnwidth]{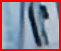}
				\hspace{-0.4cm} &
				\includegraphics[width=0.13\columnwidth]{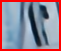}
				\hspace{-0.4cm} &
				\includegraphics[width=0.13\columnwidth]{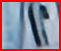}
				\hspace{-0.4cm} &
				\includegraphics[width=0.13\columnwidth]{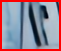} 
				\hspace{-0.4cm} &
				\includegraphics[width=0.13\columnwidth]{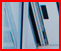} 
				\\
				\includegraphics[width=0.13\columnwidth]{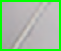}
				\hspace{-0.4cm} &
				\includegraphics[width=0.13\columnwidth]{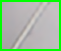}
				\hspace{-0.4cm} &
				\includegraphics[width=0.13\columnwidth]{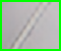}
				\hspace{-0.4cm} &
				\includegraphics[width=0.13\columnwidth]{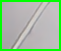} 
				\hspace{-0.4cm} &
				\includegraphics[width=0.13\columnwidth]{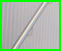} 
				\\
				(b) \hspace{-0.4cm} & 
				(c) \hspace{-0.4cm} & 
				(d) \hspace{-0.4cm} & 
				(e) \hspace{-0.4cm} & 
				(f) 
				\\ 
			\end{tabular}
		\end{adjustbox}
	\end{tabular}
	\vspace{-2mm}
	\caption{
		\textbf{Contribution of different components in LapSRN}. 
		(a) Ground truth HR image (b) without pyramid structure (c) without global residual learning (d) without robust loss (e) full model (f) HR patch.
	}
	\label{fig:ablation}
\end{figure}

By removing the pyramid structure, our model falls back to a network similar to the FSRCNN but with the global residual learning.
%
{We train this network using 10 convolutional layers in order to have the same depth as our LapSRN.}
\figref{compare_loss_curve} shows the convergence curves in terms of PSNR on the \textsc{Set14} for $4\times$ SR.
The quantitative results in~\tabref{ablation} and~\figref{ablation} show that the pyramid structure leads to considerable performance improvement (e.g., 0.7 dB on \textsc{Set5} and 0.4 dB on \textsc{Set14}), which validates the effectiveness of our Laplacian pyramid network design.

\subsubsection{Global residual learning} 
\begin{figure}
	\centering
	\includegraphics[width=0.7\linewidth]{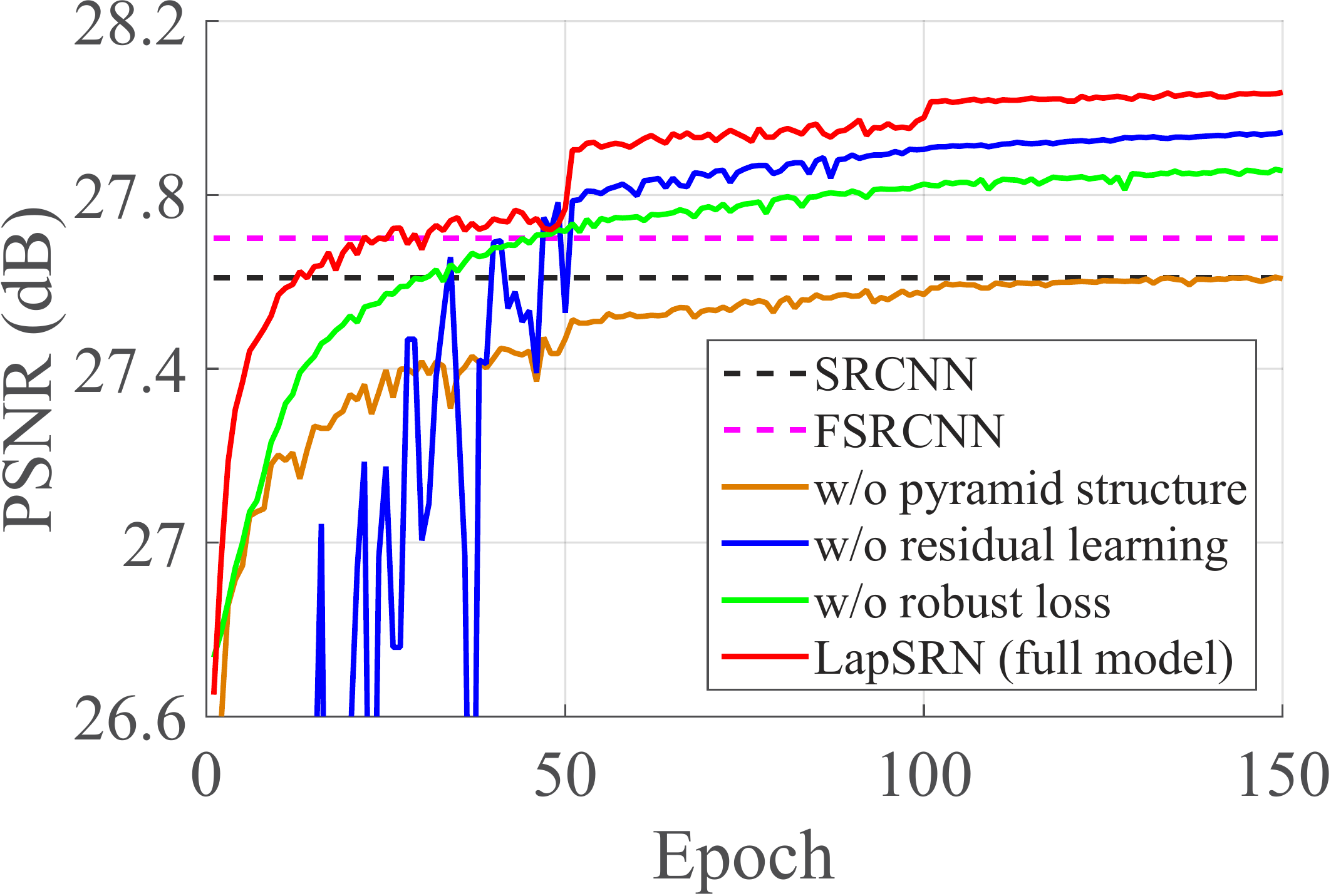} 
	\vspace{-0.3cm}
	\caption{
		\textbf{Convergence analysis}.
		We analyze the contributions of the pyramid structures, loss functions, and global residual learning by replacing each component with the one used in existing methods.
		Our full model converges faster and achieves better performance.
	}
	\label{fig:compare_loss_curve}
	\vspace{-3mm}
\end{figure}

To demonstrate the effectiveness of global residual learning, we remove the image reconstruction branch and directly predict the HR images at each level.
In~\figref{compare_loss_curve}, the performance of the non-residual network (blue curve) converges slowly and fluctuates significantly during training.
Our full LapSRN model (red curve), on the other hand, outperforms the SRCNN within 10 epochs.

\begin{table}
	\centering
	\caption{
		\textbf{Ablation study of LapSRN.}
		Our full model performs favorably against several variants of the LapSRN on both \textsc{Set5} and \textsc{Set14} for $4\times$ SR.
	}
	\vspace{-2mm}
	\begin{tabular}{ccc|cc}
		\toprule
		GRL & Pyramid & Loss & \textsc{Set5} & \textsc{Set14} \\
		\midrule
		\checkmark & & Charbonnier & 30.58 & 27.61 \\
		& \checkmark & Charbonnier & 31.10 & 27.94 \\
		\checkmark & \checkmark & $\mathcal{L}_2$ & 30.93 & 27.86 \\
		\checkmark & \checkmark & Charbonnier & \textbf{31.28} & \textbf{28.04} \\
		\bottomrule
	\end{tabular}
	\label{tab:ablation}
\end{table}

\begin{figure}
	\scriptsize
	\centering
	\begin{tabular}{cc}
		\hspace{-0.45cm}
		\begin{adjustbox}{valign=t}
			\begin{tabular}{c}
				\includegraphics[width=0.29\columnwidth]{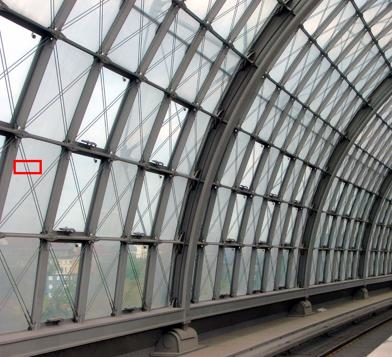}
				\\
				Ground-truth HR
			\end{tabular}
		\end{adjustbox}
		\hspace{-0.45cm}
		\begin{adjustbox}{valign=t}
			\begin{tabular}{ccc}
				\includegraphics[width=0.22\columnwidth]{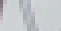}
				\hspace{-0.37cm} &
				\includegraphics[width=0.22\columnwidth]{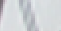}
				\hspace{-0.37cm} &
				\includegraphics[width=0.22\columnwidth]{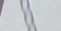}
				\\
				LR \hspace{-0.37cm} &
				$2\times$ SR w/o M.S. \hspace{-0.37cm} &
				$4\times$ SR w/o M.S. 
				\\
				\includegraphics[width=0.22\columnwidth]{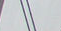}
				\hspace{-0.37cm} &								\includegraphics[width=0.22\columnwidth]{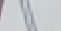}
				\hspace{-0.37cm} &								\includegraphics[width=0.22\columnwidth]{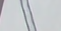}
				\\
				HR \hspace{-0.37cm} &
				$2\times$ SR w/ M.S. \hspace{-0.37cm} &
				$4\times$ SR w/ M.S. 
				\\
			\end{tabular}
		\end{adjustbox}
	\end{tabular}
	\vspace{-2mm}
	\caption{
		\textbf{Contribution of multi-scale supervision (M.S.)}. 
		The multi-scale supervision guides the network training to progressively reconstruct the HR images and help reduce the spatial aliasing artifacts.
}
	\label{fig:deep_supervision}
\end{figure}

\subsubsection{Loss function} 
To validate the effectiveness of the Charbonnier loss function, we train the proposed network with conventional $\mathcal{L}_2$ loss function.
We use a larger learning rate ($1e-4$) since the gradient magnitude of the $\mathcal{L}_2$ loss is smaller. 
As illustrated in~\figref{compare_loss_curve}, the network optimized with the $\mathcal{L}_2$ loss (green curve) requires more iterations to achieve comparable performance with SRCNN. 
In~\figref{ablation}, we show that the SR images reconstruct by our full model contain relatively clean and sharp details. 

\subsubsection{Multi-scale supervision} 
{
As described in~\secref{loss}, we use the multiple loss functions to supervise the intermediate output at each pyramid level.
%
%
We show the intermediate output images at each pyramid scale in~\figref{deep_supervision}.
The model without the multi-scale supervision (i.e., only applying supervision at the finest scale) cannot reduce the spatial aliasing artifacts well, while our LapSRN progressively reconstructs clear and sharp straight lines.
}

\subsection{Parameter sharing}
\label{sec:exp_parameter_sharing}
In this section, we reduce the network parameters in our LapSRN by sharing weights \emph{across} and \emph{within} pyramid levels and discuss the performance contribution.

\subsubsection{Parameter sharing across pyramid levels}
\label{sec:exp_parameter_sharing_between}
Our preliminary LapSRN $4\times$ model~\cite{LapSRN} has 812k parameters as each pyramid level has distinct convolutional and transposed convolutional layers.
By sharing the weights across pyramid levels as shown in~\figref{parameter_sharing}, we reduce the number of parameters to 407k.
Such model has 10 convolutional layers, 1 recursive block, and does not use any local residual learning strategies.
We denote this model by $\text{LapSRN}_{\textbf{NS}}$-D10R1.
We compare the above models on the \textsc{BSDS100} and \textsc{Urban100} datasets for $4\times$ SR.
\tabref{reduce_parameters} shows that the $\text{LapSRN}_{\textbf{NS}}$-D10R1 achieves comparable performance with the LapSRN~\cite{LapSRN} while using only half of the network parameters.

\subsubsection{Parameter sharing within pyramid levels}
\label{sec:exp_parameter_sharing_within}
We further reduce the network parameters by decreasing the number of convolutional layers (D) and increasing the number of recursive blocks (R).
We train another two models: $\text{LapSRN}_{\textbf{NS}}$-D5R2 and $\text{LapSRN}_{\textbf{NS}}$-D2R5, which have 222k and 112k parameters, respectively.
As shown in~\tabref{reduce_parameters}, while the 
$\text{LapSRN}_{\textbf{NS}}$-D5R2 and $\text{LapSRN}_{\textbf{NS}}$-D2R5 
have fewer parameters, we observe the performance drop, particularly on the challenging \textsc{Urban100} dataset.

\begin{table}
	\centering
	\caption{
		\textbf{Parameter sharing in LapSRN.}
		We reduce the number of network parameters by sharing the weights between pyramid levels and applying recursive layers in the feature embedding sub-network.
	}
	\vspace{-3mm}
	\begin{tabular}{cc|cc}
		\toprule
		Model & 
		$\#$Parameters &
		\textsc{BSDS100} &
		\textsc{Urban100} 
		\\
		\midrule
		LapSRN~\cite{LapSRN} & 812k &
		\textbf{27.32} & \textbf{25.21}
		\\
		$\text{LapSRN}_{\textbf{NS}}$-D10R1 & 407k &
		\textbf{27.32} & 25.20 
		\\
		$\text{LapSRN}_{\textbf{NS}}$-D5R2 & 222k &
		27.30 & 25.16 
		\\
		$\text{LapSRN}_{\textbf{NS}}$-D2R5 & 112k &
		27.26 & 25.10 
		\\ 
		\bottomrule
	\end{tabular}
	\label{tab:reduce_parameters}
	\vspace{-2mm}
\end{table}

\subsection{Training deeper models}
In~\secref{exp_parameter_sharing}, we show that we can achieve comparable performance to the preliminary LapSRN by using only half or $27\%$ of parameters.
Next, we train deeper models to improve the performance without increasing the number of the network parameters.

\subsubsection{Local residual learning}
\begin{figure}
	\centering
	\includegraphics[width=0.8\linewidth]{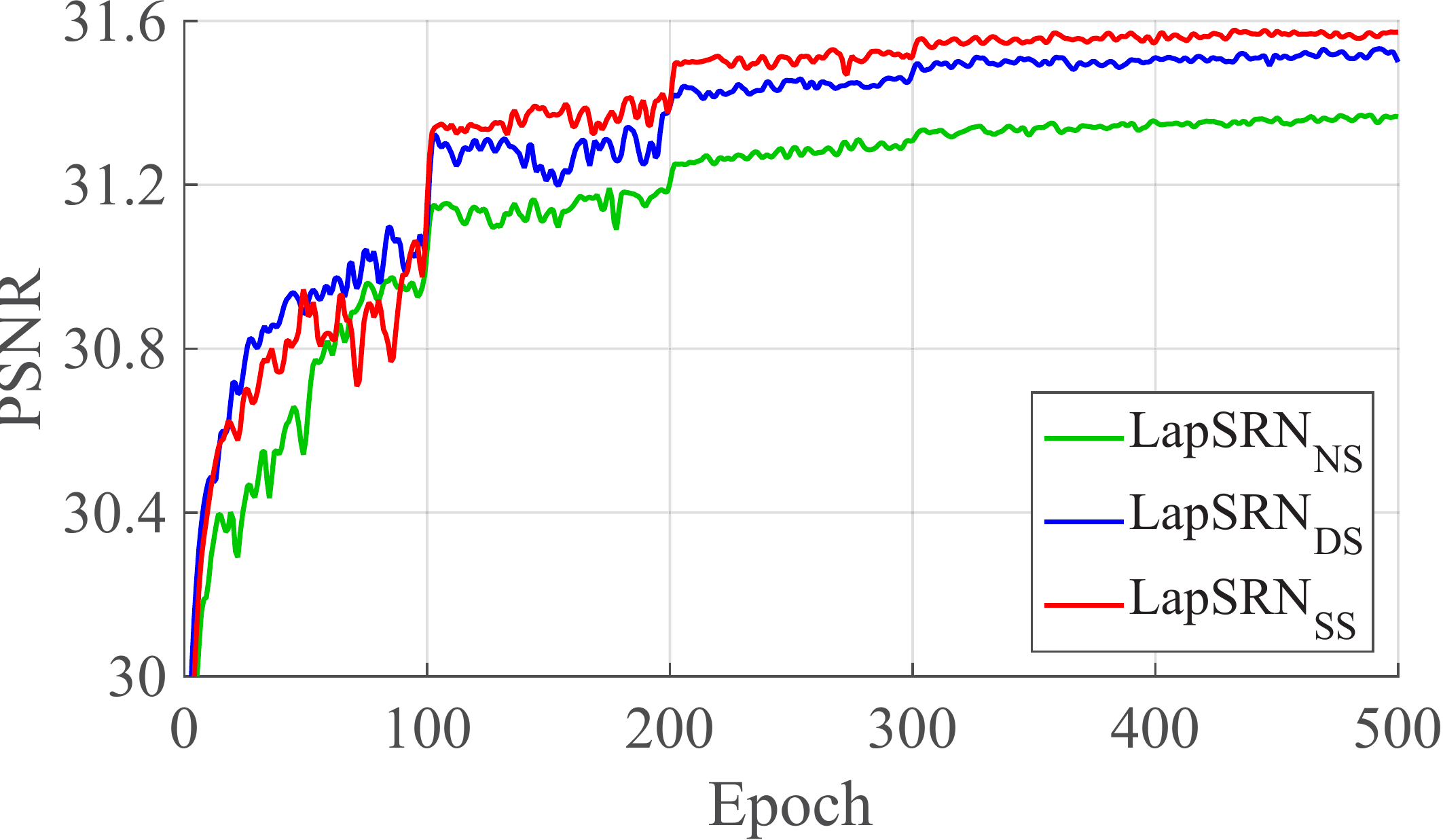}
	\vspace{-0.3cm}
	\caption{
		\textbf{Comparisons of local residual learning}. 
		We train our LapSRN-D5R5 model with three different local residual learning methods as described in~\secref{LRL} and evaluate on the \textsc{Set5} for $4\times$ SR.
	}
	\label{fig:LRL_curve}
	\vspace{-3mm}
\end{figure}

We increase the number of recursive blocks in our feature embedding sub-network to increase the depth of network but keep the number of parameters the same.
We test three LapSRN models: D5R2, D5R5, and D5R8, which have 5 distinct convolutional layers with 2, 5 and 8 recursive blocks, respectively.
%
%
%
We train the models with three different local residual learning methods as described in~\secref{LRL}.
We plot the convergence curves of the LapSRN-D5R5 in~\figref{LRL_curve} and present the quantitative evaluation in~\tabref{compare_LRL}.
Overall, the shared-source local skip connection method ($\text{LapSRN}_{\textbf{SS}}$) performs favorably against other alternatives, particularly for deeper models (i.e., more recursive blocks).

\begin{table}
	\centering
	\caption{
		\textbf{Quantitative evaluation of local residual learning}. 
		We compare three different local residual learning methods on 
		the \textsc{Urban100} dataset for $4\times$ SR.
		Overall, the shared local skip connection method ($\text{LapSRN}_{\textbf{SS}}$) achieves superior performance for deeper models.
	}
	\vspace{-2mm}
	\begin{tabular}{cc|ccc}
		\toprule
		Model & 
		Depth &
		$\text{LapSRN}_{\textbf{NS}}$ &
		$\text{LapSRN}_{\textbf{DS}}$ &
		$\text{LapSRN}_{\textbf{SS}}$ 
		\\
		\midrule
		D5R2 & 24 & 25.16 & 25.22 & \textbf{25.23} 
		\\ 
		D5R5 & 54 & 25.18 & 25.33 & \textbf{25.34}
		\\ 
		D5R8 & 84 & 25.26 & 25.33 & \first{25.38}
		\\
		\bottomrule
	\end{tabular}
	\
	\label{tab:compare_LRL}
\end{table}

\subsubsection{Study of D and R}
\begin{figure}
	\centering
	\includegraphics[width=0.85\columnwidth]{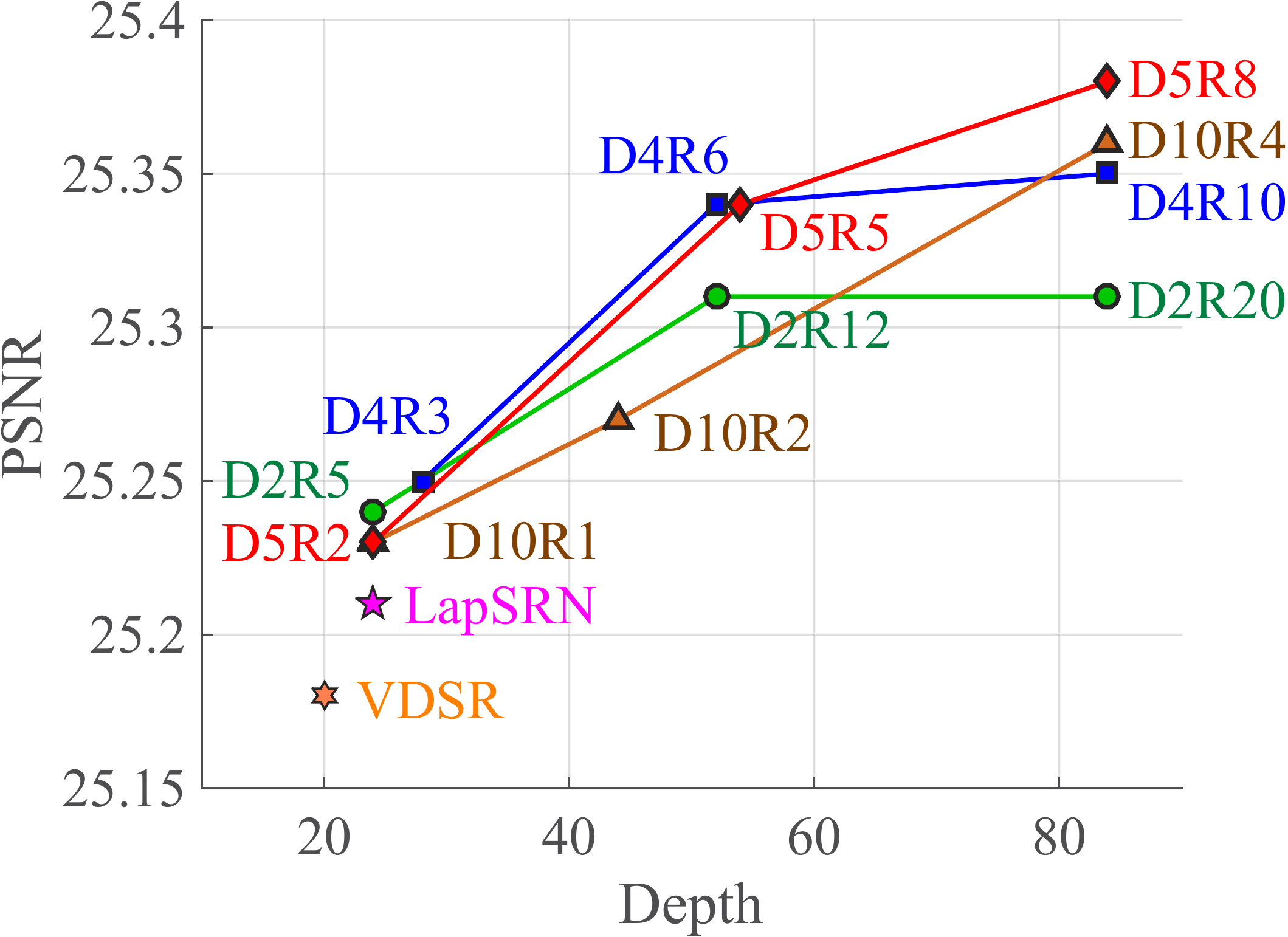}
	\vspace{-2mm}
	\caption{
		\textbf{PSNR versus network depth}.
		We test the proposed model with different $D$ and $R$ on the \textsc{Urban100}
		dataset for $4\times$ SR.
	}
	\label{fig:BD_curve}
\end{figure}

Our feature embedding sub-network consists of $R$ recursive blocks, and each recursive block has $D$ distinct convolutional layers which are shared among all the recursive blocks.
Here we extensively evaluate the contributions of R and D to the reconstruction accuracy.
We use D = $2, 4, 5, 10$ to construct models with different network depth.
We use the shared-source local skip connection for all the evaluated models.
We show the quantitative evaluation in~\tabref{study_BD} and visualize the performance over the network depth in~\figref{BD_curve}.
While the D2R5, D5R2, and D10R1 models perform comparably, 
the D5R8 method achieves the best reconstruction accuracy when the network depth is more than 80.

\begin{table}
	\centering
	\caption{
		\textbf{Quantitative evaluation of the number of recursive blocks $\mathbf{R}$ and the number of convolutional layers $\mathbf{D}$ in our feature embedding sub-network}. 
		We build LapSRN with different network depth by varying the values of D and R and evaluate on the \textsc{BSDS100} and \textsc{Urban100} datasets for $4\times$ SR.
	}
	\vspace{-1mm}
	\begin{tabular}{lcc|cc}
		\toprule
		Model &
		$\#$Parameters & 
		Depth &
		\textsc{BSDS100} &
		\textsc{Urban100}
		\\
		\midrule
		D2R5 & 112k & 24 & 27.33 & 25.24 
		\\ 
		D2R12 & 112k & 52 & 27.35 & \textbf{25.31}
		\\ 
		D2R20 & 112k & 84 & \textbf{27.37} & \textbf{25.31}
		\\
		\midrule
		D4R3 & 185k & 28 & 27.33 & 25.25 
		\\ 
		D4R6 & 185k & 52 & \textbf{27.37} & 25.34
		\\ 
		D4R10 & 185k & 84 & \textbf{27.37} & \textbf{25.35}
		\\
		\midrule
		D5R2 & 222k & 24 & 27.32 & 25.23 
		\\ 
		D5R5 & 222k & 54 & 27.38 & 25.34
		\\ 
		D5R8 & 222k & 84 & \first{27.39} & \first{25.38}
		\\
		\midrule
		D10R1 & 407k & 24 & 27.33 & 25.23 
		\\ 
		D10R2 & 407k & 44 & 27.36 & 25.27
		\\ 
		D10R4 & 407k & 84 & \textbf{27.38} & \textbf{25.36}
		\\
		\bottomrule
	\end{tabular}
	\
	\label{tab:study_BD}
\end{table}

\subsection{Multi-scale training}
We train our LapSRN using the multi-scale training strategy (\secref{multiscale}).
As our pyramid network design only accounts for training with $2^n\times$ samples, we train our $\text{LapSRN}_{\textbf{SS}}$-D5R8 model with the following scale combinations: $\{2\times\}$, $\{4\times\}$, $\{8\times\}$, $\{2\times, 4\times\}$, $\{2\times, 8\times\}$, $\{4\times, 8\times\}$ and $\{2\times, 4\times, 8\times\}$.
During training, we equally split a batch of samples for every upsampling scale.
Note that all these models have the same numbers of parameters due to parameter sharing.
We evaluate the above models for $2\times$, $4\times$ and $8\times$ SR by constructing LapSRN with the corresponding pyramid levels.
We also evaluate $3\times$ SR using our 2-level LapSRN and resizing the network output to the desired spatial resolution.
%
%
From our experimental results, the model trained with $2\times, 4\times$ and $8\times$ samples has the capacity to handle multiple upsampling scales and generalizes well to the unseen $3\times$ SR examples.
Furthermore, the multi-scale models perform favorably against the single-scale models, particularly on the \textsc{Urban100} dataset. 
%
%
Due to the space limit, we present complete quantitative and visual comparisons in the supplementary material.


\vspace{-2mm}
\section{Experiment Results}
\label{sec:experiments}

In this section, we compare the proposed LapSRN with several state-of-the-art SR methods on benchmark datasets.
We present the quantitative evaluation, qualitative comparison, runtime, and parameters comparisons.
We then evaluate our method on real-world photos, compare with the LAPGAN~\cite{LAPGAN}, and incorporate the adversarial training.
{
In addition, we conduct a human subject study using the pairwise comparison and provide the analysis in the supplementary material.
}
Finally, we discuss the limitation of the proposed method. 
We provide our source code and SR results generated by all the evaluated methods on our project website at \url{http://vllab.ucmerced.edu/wlai24/LapSRN}.

\begin{table*}
	\centering
	\caption{
		\textbf{Quantitative evaluation of state-of-the-art SR algorithms}. 
		We report the average PSNR/SSIM/IFC for $2\times$, $3\times$, $4\times$ and $8\times$ SR.
		\red{\textbf{Red}} and \blue{\underline{blue}} indicate the best and the second best performance, respectively.
		Both LapSRN~\cite{LapSRN} and the proposed MS-LapSRN do not use any $3\times$ SR images for training.
		To generate the results of $3\times$ SR, we first perform $4\times$ SR on input LR images and then downsample the output to the target resolution.
	}
	\vspace{-3mm}
	\label{tab:quality} 
	\resizebox{\textwidth}{!}{
	\begin{tabular}{rcccccc}
		\toprule
		\multirow{2}{*}{Algorithm}
		&
		\multirow{2}{*}{Scale}
		& 
		\textsc{Set5} & \textsc{Set14} & \textsc{BSDS100} & \textsc{Urban100} & \textsc{Manga109} \\
		& 
		&
		PSNR / SSIM / IFC & 
		PSNR / SSIM / IFC & 
		PSNR / SSIM / IFC & 
		PSNR / SSIM / IFC & 
		PSNR / SSIM / IFC \\
		\midrule
		Bicubic & \multirow{12}{*}{$2\times$} &
		33.69 / 0.931 / 6.166 &
		30.25 / 0.870 / 6.126 &
		29.57 / 0.844 / 5.695 &
		26.89 / 0.841 / 6.319 &
		30.86 / 0.936 / 6.214
		\\
		A+~\cite{A+} & &
		36.60 / 0.955 / 8.715 &
		32.32 / 0.906 / 8.200 &
		31.24 / 0.887 / 7.464 &
		29.25 / 0.895 / 8.440 &
		35.37 / 0.968 / 8.906
		\\
		RFL~\cite{RFL} & &
		36.59 / 0.954 / 8.741 &
		32.29 / 0.905 / 8.224 &
		31.18 / 0.885 / 7.473 &
		29.14 / 0.891 / 8.439 &
		35.12 / 0.966 / 8.921
		\\
		SelfExSR~\cite{SelfExSR} & &
		36.60 / 0.955 / 8.404 &
		32.24 / 0.904 / 8.018 &
		31.20 / 0.887 / 7.239 &
		29.55 / 0.898 / 8.414 &
		35.82 / 0.969 / 8.721
		\\
		SRCNN~\cite{SRCNN} & &
		36.72 / 0.955 / 8.166 &
		32.51 / 0.908 / 7.867 &
		31.38 / 0.889 / 7.242 &
		29.53 / 0.896 / 8.092 &
		35.76 / 0.968 / 8.471
		\\
		FSRCNN~\cite{FSRCNN} & &
		37.05 / 0.956 / 8.199 &
		32.66 / 0.909 / 7.841 &
		31.53 / 0.892 / 7.180 &
		29.88 / 0.902 / 8.131 &
		36.67 / 0.971 / 8.587
		\\
		SCN~\cite{SCN} & &
		36.58 / 0.954 / 7.358 &
		32.35 / 0.905 / 7.085 &
		31.26 / 0.885 / 6.500 &
		29.52 / 0.897 / 7.324 &
		35.51 / 0.967 / 7.601
		\\
		VDSR~\cite{VDSR} & &
		37.53 / \second{0.959} / 8.190 &
		33.05 / 0.913 / 7.878 &
		31.90 / 0.896 / 7.169 &
		30.77 / 0.914 / 8.270 &
		37.22 / \second{0.975} / 9.120
		\\
		DRCN~\cite{DRCN} & &
		37.63 / \second{0.959} / 8.326 &
		33.06 / 0.912 / 8.025 &
		31.85 / 0.895 / 7.220 &
		30.76 / 0.914 / 8.527 &
		37.63 / 0.974 / 9.541
		\\
		LapSRN~\cite{LapSRN} & &
		37.52 / \second{0.959} / 9.010 &
		33.08 / 0.913 / 8.501 &
		31.80 / 0.895 / 7.715 &
		30.41 / 0.910 / 8.907 &
		37.27 / 0.974 / 9.481
		\\
		DRRN~\cite{DRRN} & &
		\second{37.74} / \second{0.959} / 8.671 &
		33.23 / \second{0.914} / 8.320 &
		\first{32.05} / \second{0.897} / 7.613 &
		\first{31.23} / \first{0.919} / 8.917 &
		\first{37.92} / \first{0.976} / 9.268
		\\
		MS-LapSRN-D5R2 (ours) & &
		37.62 / \first{0.960} / 9.038 &
		33.13 / 0.913 / 8.539 &
		31.93 / \second{0.897} / 7.776 &
		30.82 / 0.915 / 9.081 &
		37.38 / \second{0.975} / 9.434
		\\
		MS-LapSRN-D5R5 (ours) & &
		37.72 / \first{0.960} / \second{9.265} &
		\second{33.24} / \second{0.914} / \second{8.726} &
		\second{32.00} / \first{0.898} / \second{7.906} &
		31.01 / \second{0.917} / \second{9.334} &
		37.71 / \second{0.975} / \second{9.710}
		\\
		MS-LapSRN-D5R8 (ours) & &
		\first{37.78} / \first{0.960} / \first{9.305} &
		\first{33.28} / \first{0.915} / \first{8.748} &
		\first{32.05} / \first{0.898} / \first{7.927} &
		\second{31.15} / \first{0.919} / \first{9.406} &
		\second{37.78} / \first{0.976} / \first{9.765}
		\\
		\midrule
		Bicubic & \multirow{12}{*}{$3\times$} &
		30.41 / 0.869 / 3.596 &
		27.55 / 0.775 / 3.491 &
		27.22 / 0.741 / 3.168 &
		24.47 / 0.737 / 3.661 &
		26.99 / 0.859 / 3.521
		\\
		A+~\cite{A+} & &
		32.62 / 0.909 / 4.979 &
		29.15 / 0.820 / 4.545 &
		28.31 / 0.785 / 4.028 &
		26.05 / 0.799 / 4.883 &
		29.93 / 0.912 / 4.880
		\\
		RFL~\cite{RFL} & &
		32.47 / 0.906 / 4.956 &
		29.07 / 0.818 / 4.533 &
		28.23 / 0.782 / 4.023 &
		25.88 / 0.792 / 4.781 &
		29.61 / 0.905 / 4.758
		\\
		SelfExSR~\cite{SelfExSR} & &
		32.66 / 0.910 / 4.911 &
		29.18 / 0.821 / 4.505 &
		28.30 / 0.786 / 3.923 &
		26.45 / 0.810 / 4.988 &
		27.57 / 0.821 / 2.193
		\\
		SRCNN~\cite{SRCNN} & &
		32.78 / 0.909 / 4.682 &
		29.32 / 0.823 / 4.372 &
		28.42 / 0.788 / 3.879 &
		26.25 / 0.801 / 4.630 &
		30.59 / 0.914 / 4.698
		\\
		FSRCNN~\cite{FSRCNN} & &
		33.18 / 0.914 / 4.970 &
		29.37 / 0.824 / 4.569 &
		28.53 / 0.791 / 4.061 &
		26.43 / 0.808 / 4.878 &
		31.10 / 0.921 / 4.912
		\\
		SCN~\cite{SCN} & &
		32.62 / 0.908 / 4.321 &
		29.16 / 0.818 / 4.006 &
		28.33 / 0.783 / 3.553 &
		26.21 / 0.801 / 4.253 &
		30.22 / 0.914 / 4.302
		\\
		VDSR~\cite{VDSR} & &
		33.67 / 0.921 / 5.088 &
		29.78 / 0.832 / 4.606 &
		28.83 / 0.799 / 4.043 &
		27.14 / 0.829 / 5.045 &
		32.01 / 0.934 / 5.389
		\\
		DRCN~\cite{DRCN} & &
		33.83 / 0.922 / 5.202 &
		29.77 / 0.832 / 4.686 &
		28.80 / 0.797 / 4.070 &
		27.15 / 0.828 / 5.187 &
		32.31 / 0.936 / 5.564
		\\
		LapSRN~\cite{LapSRN} & &
		33.82 / 0.922 / 5.194 &
		29.87 / 0.832 / 4.662 &
		28.82 / 0.798 / 4.057 &
		27.07 / 0.828 / 5.168 &
		32.21 / 0.935 / 5.406
		\\
		DRRN~\cite{DRRN} & &
		\second{34.03} / \first{0.924} / \first{5.397} &
		\second{29.96} / \second{0.835} / \first{4.878} &
		\first{28.95} / 0.800 / \first{4.269} &
		\first{27.53} / 0.764 / \first{5.456} &
		\first{32.74} / \first{0.939} / \first{5.659}
		\\
		MS-LapSRN-D5R2 (ours) & &
		33.88 / \second{0.923} / 5.165 &
		29.89 / 0.834 / 4.637 &
		28.87 / 0.800 / 4.040 &
		27.23 / 0.831 / 5.142 &
		32.28 / 0.936 / 5.384
		\\
		MS-LapSRN-D5R5 (ours) & &
		34.01 / \first{0.924} / 5.307 &
		\second{29.96} / \first{0.836} / 4.758 &
		28.92 / \second{0.801} / 4.127 &
		27.39 / \second{0.835} / 5.333 &
		32.60 / \second{0.938} / 5.559
		\\
		MS-LapSRN-D5R8 (ours) & &
		\first{34.06} / \first{0.924} / \second{5.390} &
		\first{29.97} / \first{0.836} / \second{4.806} &
		\second{28.93} / \first{0.802} / \second{4.154} &
		\second{27.47} / \first{0.837} / \second{5.409} &
		\second{32.68} / \first{0.939} / \second{5.621}
		\\
		\midrule
		Bicubic & \multirow{12}{*}{$4\times$} &
		28.43 / 0.811 / 2.337 &
		26.01 / 0.704 / 2.246 &
		25.97 / 0.670 / 1.993 &
		23.15 / 0.660 / 2.386 &
		24.93 / 0.790 / 2.289
		\\
		A+~\cite{A+} & &
		30.32 / 0.860 / 3.260 &
		27.34 / 0.751 / 2.961 &
		26.83 / 0.711 / 2.565 &
		24.34 / 0.721 / 3.218 &
		27.03 / 0.851 / 3.177
		\\
		RFL~\cite{RFL} & &
		30.17 / 0.855 / 3.205 &
		27.24 / 0.747 / 2.924 &
		26.76 / 0.708 / 2.538 &
		24.20 / 0.712 / 3.101 &
		26.80 / 0.841 / 3.055
		\\
		SelfExSR~\cite{SelfExSR} & &
		30.34 / 0.862 / 3.249 &
		27.41 / 0.753 / 2.952 &
		26.84 / 0.713 / 2.512 &
		24.83 / 0.740 / 3.381 &
		27.83 / 0.866 / 3.358
		\\
		SRCNN~\cite{SRCNN} & &
		30.50 / 0.863 / 2.997 &
		27.52 / 0.753 / 2.766 &
		26.91 / 0.712 / 2.412 &
		24.53 / 0.725 / 2.992 &
		27.66 / 0.859 / 3.045
		\\
		FSRCNN~\cite{FSRCNN} & &
		30.72 / 0.866 / 2.994 &
		27.61 / 0.755 / 2.722 &
		26.98 / 0.715 / 2.370 &
		24.62 / 0.728 / 2.916 &
		27.90 / 0.861 / 2.950
		\\
		SCN~\cite{SCN} & &
		30.41 / 0.863 / 2.911 &
		27.39 / 0.751 / 2.651 &
		26.88 / 0.711 / 2.309 &
		24.52 / 0.726 / 2.860 &
		27.39 / 0.857 / 2.889
		\\
		VDSR~\cite{VDSR} & &
		31.35 / 0.883 / 3.496 &
		28.02 / 0.768 / 3.071 &
		27.29 / 0.726 / 2.627 &
		25.18 / 0.754 / 3.405 &
		28.83 / 0.887 / 3.664
		\\
		DRCN~\cite{DRCN} & &
		31.54 / 0.884 / 3.502 &
		28.03 / 0.768 / 3.066 &
		27.24 / 0.725 / 2.587 &
		25.14 / 0.752 / 3.412 &
		28.98 / 0.887 / 3.674
		\\
		LapSRN~\cite{LapSRN} & &
		31.54 / 0.885 / 3.559 &
		28.19 / 0.772 / 3.147 &
		27.32 / 0.727 / 2.677 &
		25.21 / 0.756 / 3.530 &
		29.09 / 0.890 / 3.729
		\\
		DRRN~\cite{DRRN} & &
		\second{31.68} / \second{0.888} / 3.703 &
		28.21 / 0.772 / \second{3.252} &
		27.38 / 0.728 / \first{2.760} &
		25.44 / 0.764 / \second{3.700} &
		29.46 / \second{0.896} / 3.878
		\\
		MS-LapSRN-D5R2 (ours) & &
		31.62 / 0.887 / 3.585 &
		28.16 / 0.772 / 3.151 &
		27.36 / \second{0.729} / 2.684 &
		25.32 / 0.760 / 3.537 &
		29.18 / 0.892 / 3.750
		\\
		MS-LapSRN-D5R5 (ours) & &
		\first{31.74} / \second{0.888} / \second{3.705} &
		\second{28.25} / \second{0.773} / 3.238 &
		\second{27.42} / \first{0.731} / 2.737 &
		\second{25.45} / \second{0.765} / 3.674 &
		\second{29.48} / \second{0.896} / \second{3.888}
		\\
		MS-LapSRN-D5R8 (ours) & &
		\first{31.74} / \first{0.889} / \first{3.749} &
		\first{28.26} / \first{0.774} / \first{3.261} &
		\first{27.43} / \first{0.731} / \second{2.755} &
		\first{25.51} / \first{0.768} / \first{3.727} &
		\first{29.54} / \first{0.897} / \first{3.928}
		\\
		\midrule
		Bicubic & \multirow{12}{*}{$8\times$} &
		24.40 / 0.658 / 0.836 &
		23.10 / 0.566 / 0.784 &
		23.67 / 0.548 / 0.646 &
		20.74 / 0.516 / 0.858 &
		21.47 / 0.650 / 0.810
		\\
		A+~\cite{A+} & &
		25.53 / 0.693 / 1.077 &
		23.89 / 0.595 / 0.983 &
		24.21 / 0.569 / 0.797 &
		21.37 / 0.546 / 1.092 &
		22.39 / 0.681 / 1.056
		\\
		RFL~\cite{RFL} & &
		25.38 / 0.679 / 0.991 &
		23.79 / 0.587 / 0.916 &
		24.13 / 0.563 / 0.749 &
		21.27 / 0.536 / 0.992 &
		22.28 / 0.669 / 0.968
		\\
		SelfExSR~\cite{SelfExSR} & &
		25.49 / 0.703 / 1.121 &
		23.92 / 0.601 / 1.005 &
		24.19 / 0.568 / 0.773 &
		21.81 / 0.577 / 1.283 &
		22.99 / 0.719 / 1.244
		\\
		SRCNN~\cite{SRCNN} & &
		25.33 / 0.690 / 0.938 &
		23.76 / 0.591 / 0.865 &
		24.13 / 0.566 / 0.705 &
		21.29 / 0.544 / 0.947 &
		22.46 / 0.695 / 1.013
		\\
		FSRCNN~\cite{FSRCNN} & &
		25.60 / 0.697 / 1.016 &
		24.00 / 0.599 / 0.942 &
		24.31 / 0.572 / 0.767 &
		21.45 / 0.550 / 0.995 &
		22.72 / 0.692 / 1.009
		\\
		SCN~\cite{SCN} & &
		25.59 / 0.706 / 1.063 &
		24.02 / 0.603 / 0.967 &
		24.30 / 0.573 / 0.777 &
		21.52 / 0.560 / 1.074 &
		22.68 / 0.701 / 1.073
		\\
		VDSR~\cite{VDSR} & &
		25.93 / 0.724 / 1.199 &
		24.26 / 0.614 / 1.067 &
		24.49 / 0.583 / 0.859 &
		21.70 / 0.571 / 1.199 &
		23.16 / 0.725 / 1.263
		\\
		DRCN~\cite{DRCN} & &
		25.93 / 0.723 / 1.192 &
		24.25 / 0.614 / 1.057 &
		24.49 / 0.582 / 0.854 &
		21.71 / 0.571 / 1.197 &
		23.20 / 0.724 / 1.257
		\\
		LapSRN~\cite{LapSRN} & &
		26.15 / 0.738 / 1.302 &
		24.35 / 0.620 / 1.133 &
		24.54 / 0.586 / 0.893 &
		21.81 / 0.581 / 1.288 &
		23.39 / 0.735 / 1.352
		\\
		DRRN~\cite{DRRN} & &
		26.18 / 0.738 / 1.307 &
		24.42 / 0.622 / 1.127 &
		24.59 / 0.587 / 0.891 &
		21.88 / 0.583 / 1.299 &
		23.60 / 0.742 / 1.406
		\\
		MS-LapSRN-D5R2 (ours) & &
		\second{26.20} / 0.747 / 1.366 &
		\second{24.45} / \second{0.626} / 1.170 &
		\second{24.61} / 0.590 / 0.920 &
		\second{21.95} / 0.592 / 1.364 &
		23.70 / 0.751 / 1.470
		\\
		MS-LapSRN-D5R5 (ours) & &
		\first{26.34} / \second{0.752} / \second{1.414} &
		\first{24.57} / \first{0.629} / \second{1.200} &
		\first{24.65} / \second{0.591} / \second{0.938} &
		\first{22.06} / \second{0.597} / \second{1.426} &
		\second{23.85} / \second{0.756} / \second{1.538}
		\\
		MS-LapSRN-D5R8 (ours) & &
		\first{26.34} / \first{0.753} / \first{1.435} &
		\first{24.57} / \first{0.629} / \first{1.209} &
		\first{24.65} / \first{0.592} / \first{0.943} &
		\first{22.06} / \first{0.598} / \first{1.446} &
		\first{23.90} / \first{0.759} / \first{1.564}
		\\
		\bottomrule 
	\end{tabular}
	}
	\vspace{-3mm}
\end{table*}

\vspace{-2mm}
\subsection{Comparisons with state-of-the-arts} 
We compare the proposed method with 10 state-of-the-art SR algorithms, including dictionary-based methods (A+~\cite{A+} and RFL~\cite{RFL}), self-similarity based method (SelfExSR~\cite{SelfExSR}), and CNN-based methods (SRCNN~\cite{SRCNN}, FSRCNN~\cite{FSRCNN}, SCN~\cite{SCN}, VDSR~\cite{VDSR}, DRCN~\cite{DRCN}, DRRN~\cite{DRRN} and our preliminary approach~\cite{LapSRN}).
We carry out extensive experiments on five public benchmark datasets: \textsc{Set5}~\cite{Bevilacqua-BMVC-2012}, \textsc{Set14}~\cite{Zeyde-2010}, \textsc{BSDS100}~\cite{BSDS}, \textsc{Urban100}~\cite{SelfExSR} and \textsc{manga109}~\cite{manga109}.
The \textsc{Set5}, \textsc{Set14} and \textsc{BSDS100} datasets consist of natural scenes; 
the \textsc{Urban100} set contains challenging urban scenes images with details in different frequency bands;
and the \textsc{manga109} is a dataset of Japanese manga.

We evaluate the SR results with three widely used image quality metrics: PSNR, SSIM~\cite{SSIM}, and IFC~\cite{IFC} and compare performance on $2\times$, $3\times$, $4\times$ and $8\times$ SR.
We re-train existing methods for $8\times$ SR using the source code (A+~\cite{A+}, RFL~\cite{RFL}, SRCNN~\cite{SRCNN}, FSRCNN~\cite{FSRCNN}, VDSR~\cite{VDSR}, and DRRN~\cite{DRRN}) or our own implementation (DRCN).
Both the SelfExSR and SCN methods can naturally handle different scale factors using progressive reconstruction.
%
We use $2\times, 4\times$ and $8\times$ SR samples for training VDSR~\cite{VDSR} and DRRN~\cite{DRRN} while use only $8\times$ SR samples for other algorithms to follow the training strategies of individual methods.

We compare three variations of the proposed method: 
(1) $\text{LapSRN}_{\textbf{SS}}$-D5R2, which has similar depth as the VDSR~\cite{VDSR}, DRCN~\cite{DRCN} and LapSRN~\cite{LapSRN}, 
(2) $\text{LapSRN}_{\textbf{SS}}$-D5R5, which has the same depth as in the DRRN~\cite{DRRN}, and 
(3) $\text{LapSRN}_{\textbf{SS}}$-D5R8, which has 84 layers for $4\times$ SR.
We train the above three models using the multi-scale training strategy with $2\times, 4\times$ and $8\times$ SR samples and denote our multi-scale models as MS-LapSRN.

We show the quantitative results in~\tabref{quality}.
Our LapSRN performs favorably against existing methods especially on $4\times$ and $8\times$ SR.
In particular, our algorithm achieves higher IFC values, which has been shown to be correlated well with human perception of image super-resolution~\cite{Yang-ECCV-2014}.
We note that our method does not use any $3\times$ SR samples for training but still generates comparable results as the DRRN.

\begin{figure*}
	\newlength\fs
	\setlength{\fs}{-0.4cm}
	\scriptsize
	\centering
	\begin{tabular}{cc}
	\hspace{-0.4cm}
	\begin{adjustbox}{valign=t}
	\begin{tabular}{c}
	\includegraphics[width=0.20\textwidth]{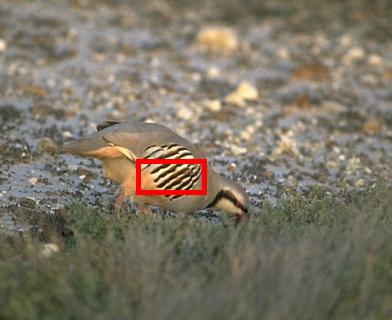}
	\\
	Ground-truth HR
	\\
	\textsc{BSDS100}: 8023
	\end{tabular}
	\end{adjustbox}
	\hspace{-0.46cm}
	\begin{adjustbox}{valign=t}
	\begin{tabular}{cccccc}
	\includegraphics[width=0.125\textwidth]{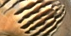} \hspace{\fs} &
	\includegraphics[width=0.125\textwidth]{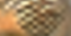} \hspace{\fs} &
	\includegraphics[width=0.125\textwidth]{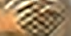} \hspace{\fs} &
	\includegraphics[width=0.125\textwidth]{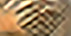} \hspace{\fs} &
	\includegraphics[width=0.125\textwidth]{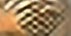} \hspace{\fs} &
	\includegraphics[width=0.125\textwidth]{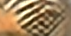}
	\\
	HR \hspace{\fs} &
	Bicubic \hspace{\fs} &
	A+~\cite{A+} \hspace{\fs} &
	SelfExSR~\cite{SelfExSR} \hspace{\fs} &
	SRCNN~\cite{SRCNN} \hspace{\fs} &
	FSRCNN~\cite{FSRCNN}
	\\
	(PSNR, SSIM, IFC) \hspace{\fs} &
	(28.50, 0.834, 2.645) \hspace{\fs} &
	(29.41, 0.860, 3.292) \hspace{\fs} &
	(29.30, 0.857, 3.159) \hspace{\fs} &
	(29.41, 0.860, 3.032) \hspace{\fs} &
	(29.83, 0.863, 3.033)
	\\
	\includegraphics[width=0.125\textwidth]{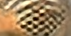} \hspace{\fs} &
	\includegraphics[width=0.125\textwidth]{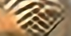} \hspace{\fs} &
	\includegraphics[width=0.125\textwidth]{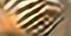} \hspace{\fs} &
	\includegraphics[width=0.125\textwidth]{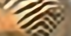} \hspace{\fs} &
	\includegraphics[width=0.125\textwidth]{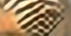} \hspace{\fs} &
	\includegraphics[width=0.125\textwidth]{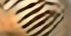} 
	\\ 
	SCN~\cite{SCN} \hspace{\fs} &
	VDSR~\cite{VDSR} \hspace{\fs} &
	DRCN~\cite{DRCN} \hspace{\fs} &
	DRRN~\cite{DRRN} \hspace{\fs} &
	LapSRN~\cite{LapSRN} \hspace{\fs} &
	MS-LapSRN (ours) 
	\\
	(29.32, 0.857, 2.889) \hspace{\fs} &
	(29.54, 0.868, 3.358) \hspace{\fs} &
	(\blue{\underline{30.29}}, 0.868, 3.302) \hspace{\fs} &
	(30.26, \blue{\underline{0.873}}, \blue{\underline{3.537}}) \hspace{\fs} &
	(30.10, 0.871, 3.432) \hspace{\fs} &
	(\red{\textbf{30.58}}, \red{\textbf{0.876}}, \red{\textbf{3.626}})
	\\
	\end{tabular}
	\end{adjustbox}
	\vspace{2mm}
	\\
	\hspace{-0.4cm}
	\begin{adjustbox}{valign=t}
	\begin{tabular}{c}
	\includegraphics[width=0.20\textwidth]{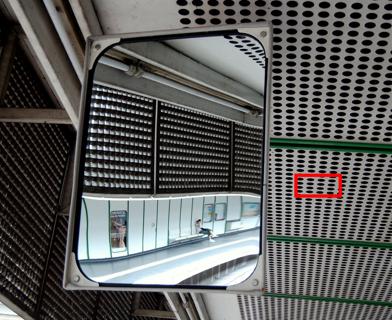}
	\\
	Ground-truth HR
	\\
	\textsc{Urban100}: img004
	\end{tabular}
	\end{adjustbox}
	\hspace{-0.46cm}
	\begin{adjustbox}{valign=t}
	\begin{tabular}{cccccc}
	\includegraphics[width=0.125\textwidth]{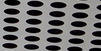} \hspace{\fs} &
	\includegraphics[width=0.125\textwidth]{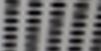} \hspace{\fs} &
	\includegraphics[width=0.125\textwidth]{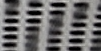} \hspace{\fs} &
	\includegraphics[width=0.125\textwidth]{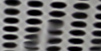} \hspace{\fs} &
	\includegraphics[width=0.125\textwidth]{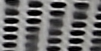} \hspace{\fs} &
	\includegraphics[width=0.125\textwidth]{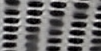}
	\\
	HR \hspace{\fs} &
	Bicubic \hspace{\fs} &
	A+~\cite{A+} \hspace{\fs} &
	SelfExSR~\cite{SelfExSR} \hspace{\fs} &
	SRCNN~\cite{SRCNN} \hspace{\fs} &
	FSRCNN~\cite{FSRCNN}
	\\
	(PSNR, SSIM, IFC) \hspace{\fs} &
	(21.11, 0.682, 2.916) \hspace{\fs} & 
	(21.96, 0.750, 3.671) \hspace{\fs} & 
	(22.68, 0.795, 4.250) \hspace{\fs} & 
	(22.17, 0.767, 3.647) \hspace{\fs} & 
	(22.03, 0.763, 3.512)
	\\
	\includegraphics[width=0.125\textwidth]{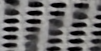} \hspace{\fs} &
	\includegraphics[width=0.125\textwidth]{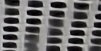} \hspace{\fs} &
	\includegraphics[width=0.125\textwidth]{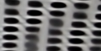} \hspace{\fs} &
	\includegraphics[width=0.125\textwidth]{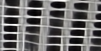} \hspace{\fs} &
	\includegraphics[width=0.125\textwidth]{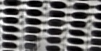} \hspace{\fs} &
	\includegraphics[width=0.125\textwidth]{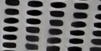} 
	\\ 
	SCN~\cite{SCN} \hspace{\fs} &
	VDSR~\cite{VDSR} \hspace{\fs} &
	DRCN~\cite{DRCN} \hspace{\fs} &
	DRRN~\cite{DRRN} \hspace{\fs} &
	LapSRN~\cite{LapSRN} \hspace{\fs} &
	MS-LapSRN (ours) 
	\\
	(22.34, 0.769, 3.601) \hspace{\fs} & 
	(22.42, 0.795, 4.309) \hspace{\fs} & 
	(\blue{\underline{22.68}}, 0.801, 4.502) \hspace{\fs} & 
	(22.35, \blue{\underline{0.802}}, \blue{\underline{4.778}}) \hspace{\fs} & 
	(22.41, 0.799, 4.527) \hspace{\fs} & 
	(\red{\textbf{22.96}}, \red{\textbf{0.822}}, \red{\textbf{4.923}})
	\\
	\end{tabular}
	\end{adjustbox}
	\vspace{2mm}
	\\
	\hspace{-0.4cm}
	\begin{adjustbox}{valign=t}
	\begin{tabular}{c}
	\includegraphics[width=0.20\textwidth]{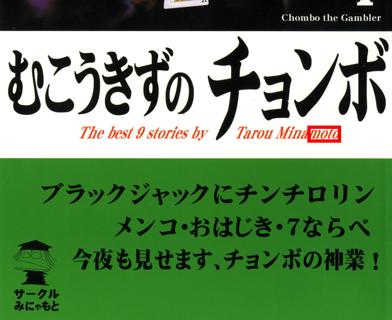}
	\\
	Ground-truth HR
	\\
	\textsc{Manga109}: MukoukizuNoCho
	\end{tabular}
	\end{adjustbox}
	\hspace{-0.46cm}
	\begin{adjustbox}{valign=t}
	\begin{tabular}{cccccc}
	\includegraphics[width=0.125\textwidth]{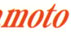} \hspace{\fs} &
	\includegraphics[width=0.125\textwidth]{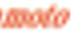} \hspace{\fs} &
	\includegraphics[width=0.125\textwidth]{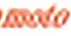} \hspace{\fs} &
	\includegraphics[width=0.125\textwidth]{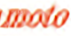} \hspace{\fs} &
	\includegraphics[width=0.125\textwidth]{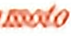} \hspace{\fs} &
	\includegraphics[width=0.125\textwidth]{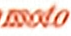}
	\\
	HR \hspace{\fs} &
	Bicubic \hspace{\fs} &
	A+~\cite{A+} \hspace{\fs} &
	SelfExSR~\cite{SelfExSR} \hspace{\fs} &
	SRCNN~\cite{SRCNN} \hspace{\fs} &
	FSRCNN~\cite{FSRCNN}
	\\
	(PSNR, SSIM, IFC) \hspace{\fs} &
	(26.28, 0.903, 2.240) \hspace{\fs} & 
	(29.67, 0.944, 3.159) \hspace{\fs} & 
	(30.77, 0.953, 3.127) \hspace{\fs} & 
	(30.96, 0.950, 3.125) \hspace{\fs} & 
	(30.89, 0.946, 2.903
	\\
	\includegraphics[width=0.125\textwidth]{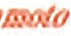} \hspace{\fs} &
	\includegraphics[width=0.125\textwidth]{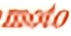} \hspace{\fs} &
	\includegraphics[width=0.125\textwidth]{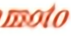} \hspace{\fs} &
	\includegraphics[width=0.125\textwidth]{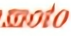} \hspace{\fs} &
	\includegraphics[width=0.125\textwidth]{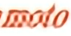} \hspace{\fs} &
	\includegraphics[width=0.125\textwidth]{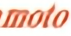} 
	\\ 
	SCN~\cite{SCN} \hspace{\fs} &
	VDSR~\cite{VDSR} \hspace{\fs} &
	DRCN~\cite{DRCN} \hspace{\fs} &
	DRRN~\cite{DRRN} \hspace{\fs} &
	LapSRN~\cite{LapSRN} \hspace{\fs} &
	MS-LapSRN (ours) 
	\\
	(30.19, 0.948, 2.925) \hspace{\fs} & 
	(32.15, 0.965, 3.744) \hspace{\fs} & 
	(32.32, 0.965, 3.676) \hspace{\fs} & 
	(\blue{\underline{32.75}}, \blue{\underline{0.968}}, \blue{\underline{3.886}}) \hspace{\fs} & 
	(32.41, 0.967, 3.821) \hspace{\fs} & 
	(\red{\textbf{32.99}}, \red{\textbf{0.969}}, \red{\textbf{4.071}})
	\\
	\end{tabular}
	\end{adjustbox}
	\end{tabular}
	\caption{
		\textbf{Visual comparison for $4\times$ SR on the \textsc{BSDS100}, \textsc{Urban100} and \textsc{Manga109} datasets.}
	}
	\label{fig:result_4x}
\end{figure*}

\begin{figure*}
	\setlength{\fs}{-0.4cm}
	\scriptsize
	\centering
	\begin{tabular}{cc}
	\hspace{-0.4cm}
	\begin{adjustbox}{valign=t}
	\begin{tabular}{c}
	\includegraphics[width=0.20\textwidth]{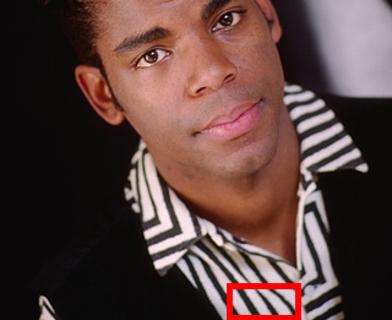}
	\\
	Ground-truth HR
	\\
	\textsc{BSDS100}: 302008
	\end{tabular}
	\end{adjustbox}
	\hspace{-0.46cm}
	\begin{adjustbox}{valign=t}
	\begin{tabular}{cccccc}
	\includegraphics[width=0.125\textwidth]{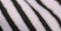} \hspace{\fs} &
	\includegraphics[width=0.125\textwidth]{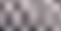} \hspace{\fs} &
	\includegraphics[width=0.125\textwidth]{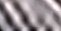} \hspace{\fs} &
	\includegraphics[width=0.125\textwidth]{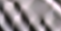} \hspace{\fs} &
	\includegraphics[width=0.125\textwidth]{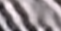} \hspace{\fs} &
	\includegraphics[width=0.125\textwidth]{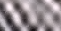}
	\\
	HR \hspace{\fs} &
	Bicubic \hspace{\fs} &
	A+~\cite{A+} \hspace{\fs} &
	SelfExSR~\cite{SelfExSR} \hspace{\fs} &
	SRCNN~\cite{SRCNN} \hspace{\fs} &
	FSRCNN~\cite{FSRCNN}
	\\
	(PSNR, SSIM, IFC) \hspace{\fs} &
	(21.45, 0.764, 0.932) \hspace{\fs} & 
	(22.34, 0.793, 1.138) \hspace{\fs} & 
	(22.32, 0.802, 1.099) \hspace{\fs} & 
	(22.46, 0.797, 0.927) \hspace{\fs} & 
	(22.34, 0.785, 0.949)
	\\
	\includegraphics[width=0.125\textwidth]{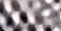} \hspace{\fs} &
	\includegraphics[width=0.125\textwidth]{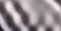} \hspace{\fs} &
	\includegraphics[width=0.125\textwidth]{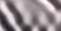} \hspace{\fs} &
	\includegraphics[width=0.125\textwidth]{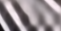} \hspace{\fs} &
	\includegraphics[width=0.125\textwidth]{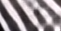} \hspace{\fs} &
	\includegraphics[width=0.125\textwidth]{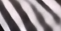} 
	\\ 
	SCN~\cite{SCN} \hspace{\fs} &
	VDSR~\cite{VDSR} \hspace{\fs} &
	DRCN~\cite{DRCN} \hspace{\fs} &
	DRRN~\cite{DRRN} \hspace{\fs} &
	LapSRN~\cite{LapSRN} \hspace{\fs} &
	MS-LapSRN (ours) 
	\\
	(22.33, 0.801, 1.043) \hspace{\fs} & 
	(22.68, 0.818, 1.195) \hspace{\fs} & 
	(22.71, 0.819, 1.210) \hspace{\fs} & 
	(\blue{\underline{22.77}}, 0.824, 1.238) \hspace{\fs} & 
	(22.52, \blue{\underline{0.829}}, \blue{\underline{1.305}}) \hspace{\fs} & 
	(\red{\textbf{22.99}}, \red{\textbf{0.840}}, \red{\textbf{1.460}})
	\\
	\end{tabular}
	\end{adjustbox}
	\vspace{2mm}
	\\ 
	\hspace{-0.4cm}
	\begin{adjustbox}{valign=t}
	\begin{tabular}{c}
	\includegraphics[width=0.20\textwidth]{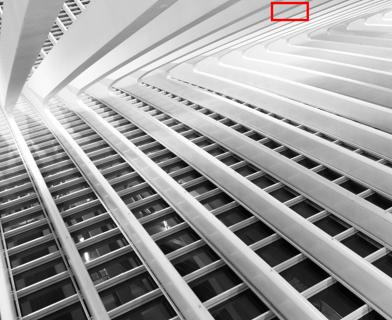}
	\\
	Ground-truth HR
	\\
	\textsc{Urban100}: img-042
	\end{tabular}
	\end{adjustbox}
	\hspace{-0.46cm}
	\begin{adjustbox}{valign=t}
	\begin{tabular}{cccccc}
	\includegraphics[width=0.125\textwidth]{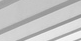} \hspace{\fs} &
	\includegraphics[width=0.125\textwidth]{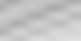} \hspace{\fs} &
	\includegraphics[width=0.125\textwidth]{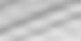} \hspace{\fs} &
	\includegraphics[width=0.125\textwidth]{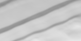} \hspace{\fs} &
	\includegraphics[width=0.125\textwidth]{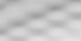} \hspace{\fs} &
	\includegraphics[width=0.125\textwidth]{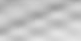}
	\\
	HR \hspace{\fs} &
	Bicubic \hspace{\fs} &
	A+~\cite{A+} \hspace{\fs} &
	SelfExSR~\cite{SelfExSR} \hspace{\fs} &
	SRCNN~\cite{SRCNN} \hspace{\fs} &
	FSRCNN~\cite{FSRCNN}
	\\
	(PSNR, SSIM, IFC) \hspace{\fs} &
	(21.67, 0.624, 1.160) \hspace{\fs} & 
	(22.66, 0.666, 1.580) \hspace{\fs} & 
	(23.18, 0.684, 1.831) \hspace{\fs} & 
	(22.56, 0.658, 1.269) \hspace{\fs} & 
	(22.52, 0.648, 1.335)
	\\
	\includegraphics[width=0.125\textwidth]{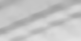} \hspace{\fs} &
	\includegraphics[width=0.125\textwidth]{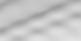} \hspace{\fs} &
	\includegraphics[width=0.125\textwidth]{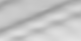} \hspace{\fs} &
	\includegraphics[width=0.125\textwidth]{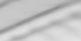} \hspace{\fs} &
	\includegraphics[width=0.125\textwidth]{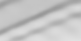} \hspace{\fs} &
	\includegraphics[width=0.125\textwidth]{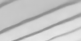} 
	\\ 
	SCN~\cite{SCN} \hspace{\fs} &
	VDSR~\cite{VDSR} \hspace{\fs} &
	DRCN~\cite{DRCN} \hspace{\fs} &
	DRRN~\cite{DRRN} \hspace{\fs} &
	LapSRN~\cite{LapSRN} \hspace{\fs} &
	MS-LapSRN (ours) 
	\\
	(22.85, 0.676, 1.510) \hspace{\fs} & 
	(23.09, 0.698, 1.738) \hspace{\fs} & 
	(23.08, 0.694, 1.699) \hspace{\fs} & 
	(\blue{\underline{23.32}}, 0.708, \blue{\underline{1.899}}) \hspace{\fs} & 
	(23.24, \blue{\underline{0.709}}, 1.887) \hspace{\fs} & 
	(\red{\textbf{23.46}}, \red{\textbf{0.726}}, \red{\textbf{2.168}})
	\\
	\end{tabular}
	\end{adjustbox}
	\vspace{2mm}
	\\
	\hspace{-0.4cm}
	\begin{adjustbox}{valign=t}
	\begin{tabular}{c}
	\includegraphics[width=0.20\textwidth]{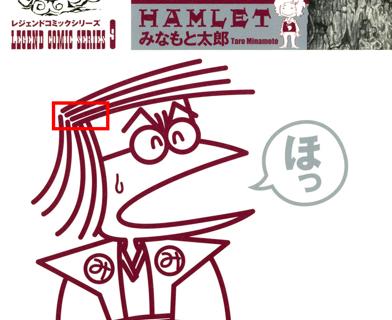}
	\\
	Ground-truth HR
	\\
	\textsc{Manga109}: Hamlet
	\end{tabular}
	\end{adjustbox}
	\hspace{-0.46cm}
	\begin{adjustbox}{valign=t}
	\begin{tabular}{cccccc}
	\includegraphics[width=0.125\textwidth]{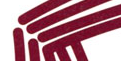} \hspace{\fs} &
	\includegraphics[width=0.125\textwidth]{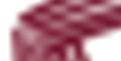} \hspace{\fs} &
	\includegraphics[width=0.125\textwidth]{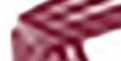} \hspace{\fs} &
	\includegraphics[width=0.125\textwidth]{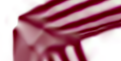} \hspace{\fs} &
	\includegraphics[width=0.125\textwidth]{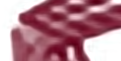} \hspace{\fs} &
	\includegraphics[width=0.125\textwidth]{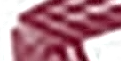}
	\\
	HR \hspace{\fs} &
	Bicubic \hspace{\fs} &
	A+~\cite{A+} \hspace{\fs} &
	SelfExSR~\cite{SelfExSR} \hspace{\fs} &
	SRCNN~\cite{SRCNN} \hspace{\fs} &
	FSRCNN~\cite{FSRCNN}
	\\
	(PSNR, SSIM, IFC) \hspace{\fs} &
	(18.36, 0.687, 0.859) \hspace{\fs} & 
	(19.64, 0.736, 1.105) \hspace{\fs} & 
	(20.41, 0.789, 1.324) \hspace{\fs} & 
	(19.89, 0.776, 1.175) \hspace{\fs} & 
	(19.83, 0.737, 1.028)
	\\
	\includegraphics[width=0.125\textwidth]{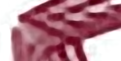} \hspace{\fs} &
	\includegraphics[width=0.125\textwidth]{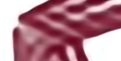} \hspace{\fs} &
	\includegraphics[width=0.125\textwidth]{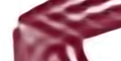} \hspace{\fs} &
	\includegraphics[width=0.125\textwidth]{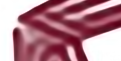} \hspace{\fs} &
	\includegraphics[width=0.125\textwidth]{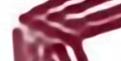} \hspace{\fs} &
	\includegraphics[width=0.125\textwidth]{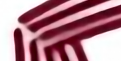} 
	\\ 
	SCN~\cite{SCN} \hspace{\fs} &
	VDSR~\cite{VDSR} \hspace{\fs} &
	DRCN~\cite{DRCN} \hspace{\fs} &
	DRRN~\cite{DRRN} \hspace{\fs} &
	LapSRN~\cite{LapSRN} \hspace{\fs} &
	MS-LapSRN (ours) 
	\\
	(19.97, 0.770, 1.170) \hspace{\fs} & 
	(21.12, 0.828, 1.668) \hspace{\fs} & 
	(21.24, 0.827, 1.680) \hspace{\fs} & 
	(\blue{\underline{21.72}}, \blue{\underline{0.848}}, \blue{\underline{1.953}}) \hspace{\fs} & 
	(21.36, 0.838, 1.731) \hspace{\fs} & 
	(\red{\textbf{21.94}}, \red{\textbf{0.862}}, \red{\textbf{2.124}})
	\\
	\end{tabular}
	\end{adjustbox}
	\end{tabular}
	\caption{
		\textbf{Visual comparison for $8\times$ SR on the \textsc{BSDS100}, \textsc{Urban100} and \textsc{Manga109} datasets.}
	}
	\label{fig:result_8x}
\end{figure*}

We show visual comparisons on the \textsc{BSDS100}, \textsc{Urban100} and \textsc{Manga109} datasets for $4\times$ SR in~\figref{result_4x} and $8\times$ SR in~\figref{result_8x}.
Our method accurately reconstructs parallel straight lines, grid patterns, and texts.
We observe that the results generated from pre-upsampling based methods~\cite{SRCNN,VDSR,DRRN} still contain noticeable artifacts caused by spatial aliasing.
In contrast, our approach effectively suppresses such artifacts through progressive reconstruction and the robust loss function.
For $8\times$ SR, it is challenging to predict HR images from bicubic-upsampled input~\cite{SRCNN,VDSR,A+} or using one-step upsampling~\cite{FSRCNN}.
The state-of-the-art methods do not super-resolve the fine structures well.
In contrast, our MS-LapSRN reconstructs high-quality HR images at a relatively fast speed.

We note that the direct reconstruction based methods, e.g., VDSR~\cite{VDSR}, can also upsample LR images progressively by iteratively applying the $2\times$ SR model.
We present detailed comparisons and analysis of such a progressive reconstruction strategy in the supplementary material.

\vspace{-1mm}
\subsection{Execution time}
\label{sec:time}
\begin{figure}
	\centering
	\includegraphics[width=0.95\columnwidth]{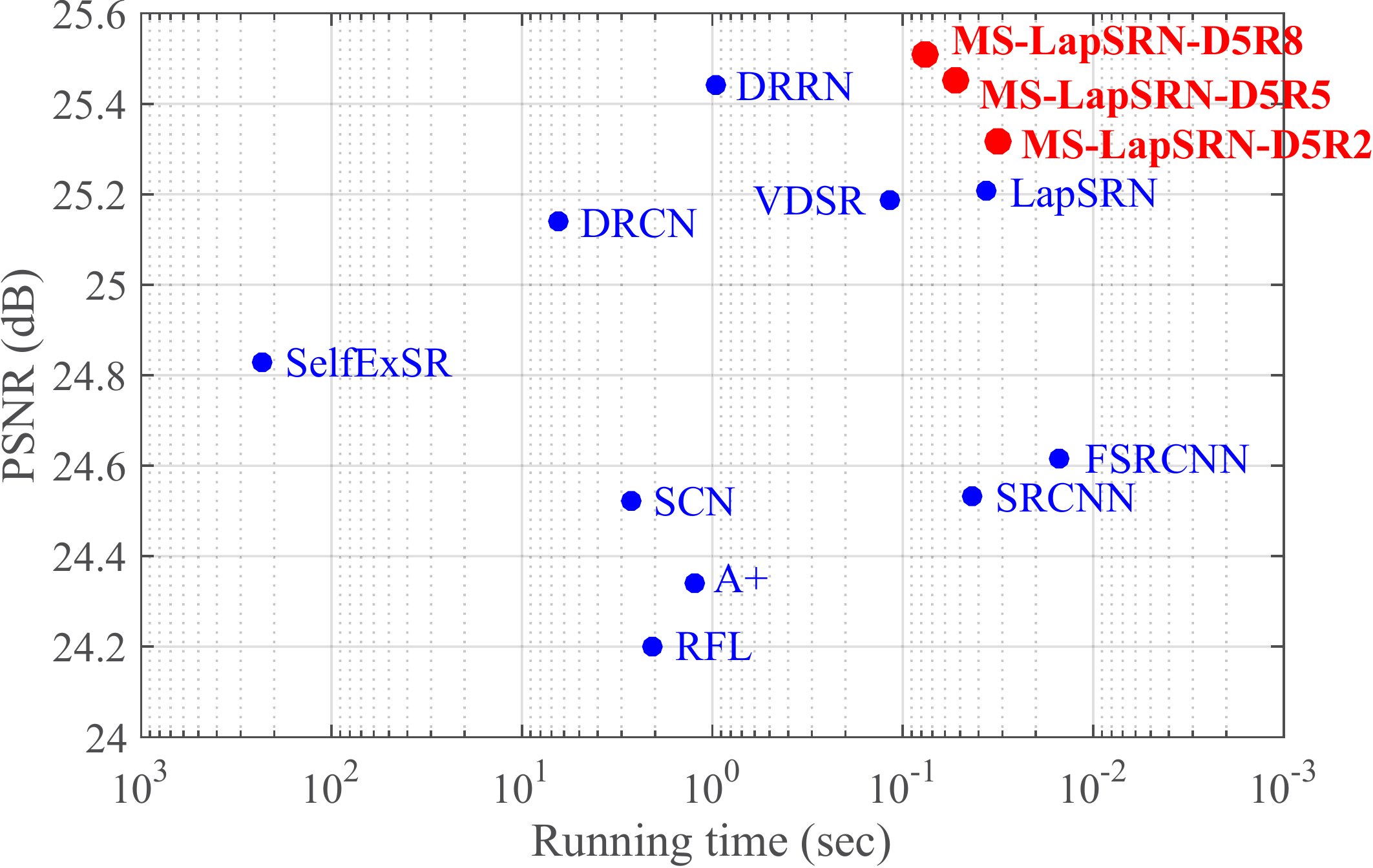}
	\caption{
		\textbf{Runtime versus performance}.
		The results are evaluated on the \textsc{Urban100} dataset for $4\times$ SR. 
		The proposed MS-LapSRN strides a balance between reconstruction accuracy and execution time.
	}
	\label{fig:psnr_time}
\end{figure}

We use the source codes of state-of-the-art methods to evaluate the runtime on the same machine with 3.4 GHz Intel i7 CPU (32G RAM) and NVIDIA Titan Xp GPU (12G Memory).
Since the testing code of the SRCNN~\cite{SRCNN} and FSRCNN~\cite{FSRCNN} is based on CPU implementation, we rebuild these models in MatConvNet to measure the runtime on GPU.
\figref{psnr_time} shows the trade-offs between the runtime and performance (in terms of PSNR) on the \textsc{Urban100} dataset for $4\times$ SR.
The speed of our MS-LapSRN-D5R2 is faster than all the existing methods except the FSRCNN~\cite{FSRCNN}.
Our MS-LapSRN-D5R8 model outperforms the state-of-the-art DRRN~\cite{DRRN} method and is an order of magnitude faster.

\begin{figure}
	\centering
	\includegraphics[width=0.95\columnwidth]{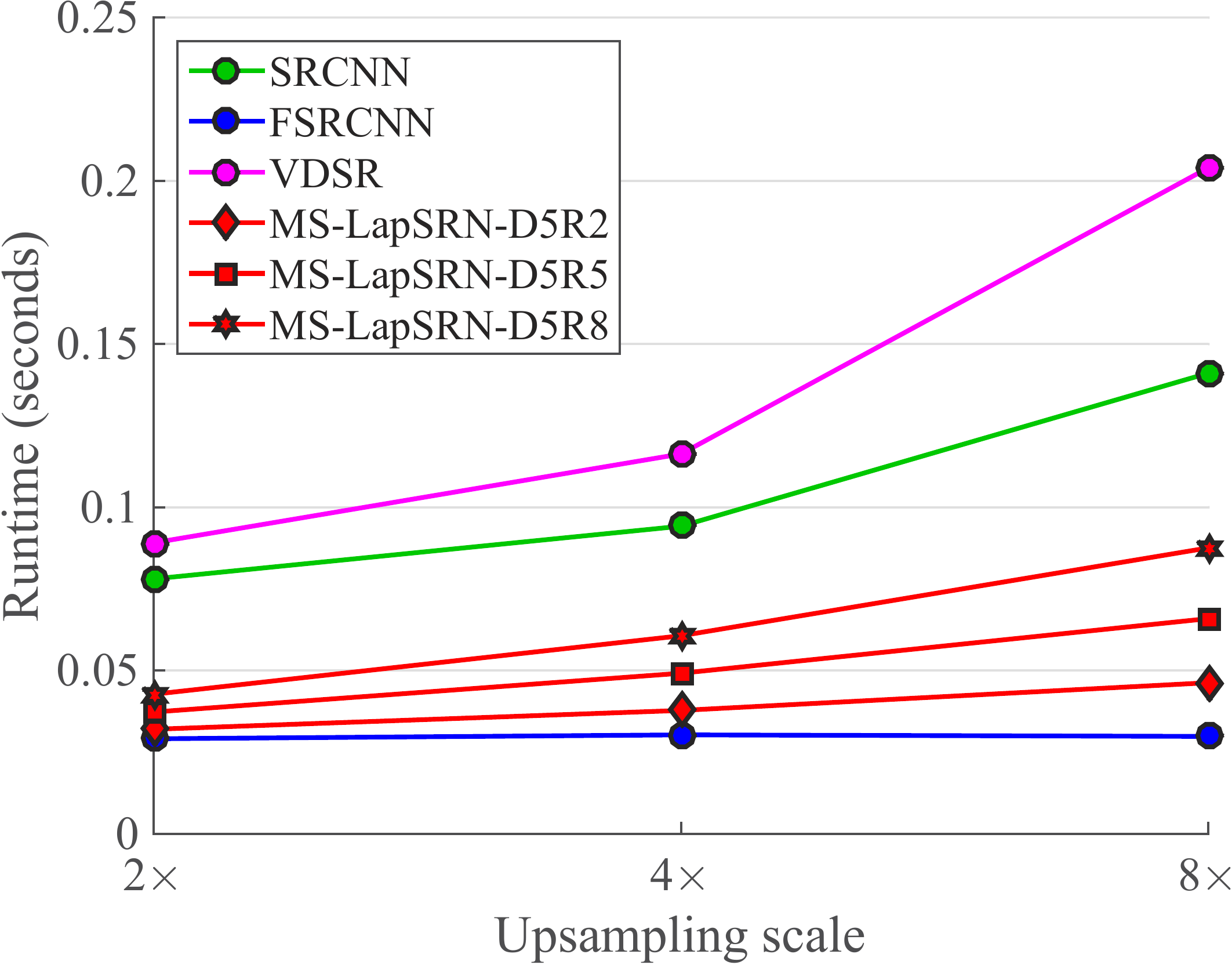}
	\caption{
		\textbf{Trade-off between runtime and upsampling scales}. 
		We fix the size of input images to $128 \times 128$ and perform $2\times$, $4\times$ and $8\times$ SR with the SRCNN~\cite{SRCNN}, FSRCNN~\cite{FSRCNN}, VDSR~\cite{VDSR} and three variations of MS-LapSRN, respectively.
	}
	\label{fig:size_time}
\end{figure}

Next, we focus on comparisons between fast CNN-based methods: SRCNN~\cite{SRCNN}, FSRCNN~\cite{FSRCNN}, VDSR~\cite{VDSR}, and LapSRN~\cite{LapSRN}.
We take an LR image with a spatial resolution of $128 \times 128$, and perform $2\times$, $4\times$ and $8\times$ SR, respectively.
We evaluate each method for 10 times and report the averaged runtime in~\figref{size_time}.
The FSRCNN is the fastest algorithm since it applies convolution on LR images and has less number of convolutional layers and filters.
The runtime of the SRCNN and VDSR depends on the size of \emph{output} images, while the speed of the FSRCNN and LapSRN is mainly determined by the size of \emph{input} images.
As the proposed LapSRN progressively upscales images and applies more convolutional layers for larger upsampling scales (i.e., require more pyramid levels), the time complexity slightly increases with respect to the desired upsampling scales.
However, the speed of our LapSRN still performs favorably against the SRCNN, VDSR, and other existing methods.

\subsection{Model parameters}
\label{sec:parameter}

\begin{figure}
	\centering
	\includegraphics[width=0.95\columnwidth]{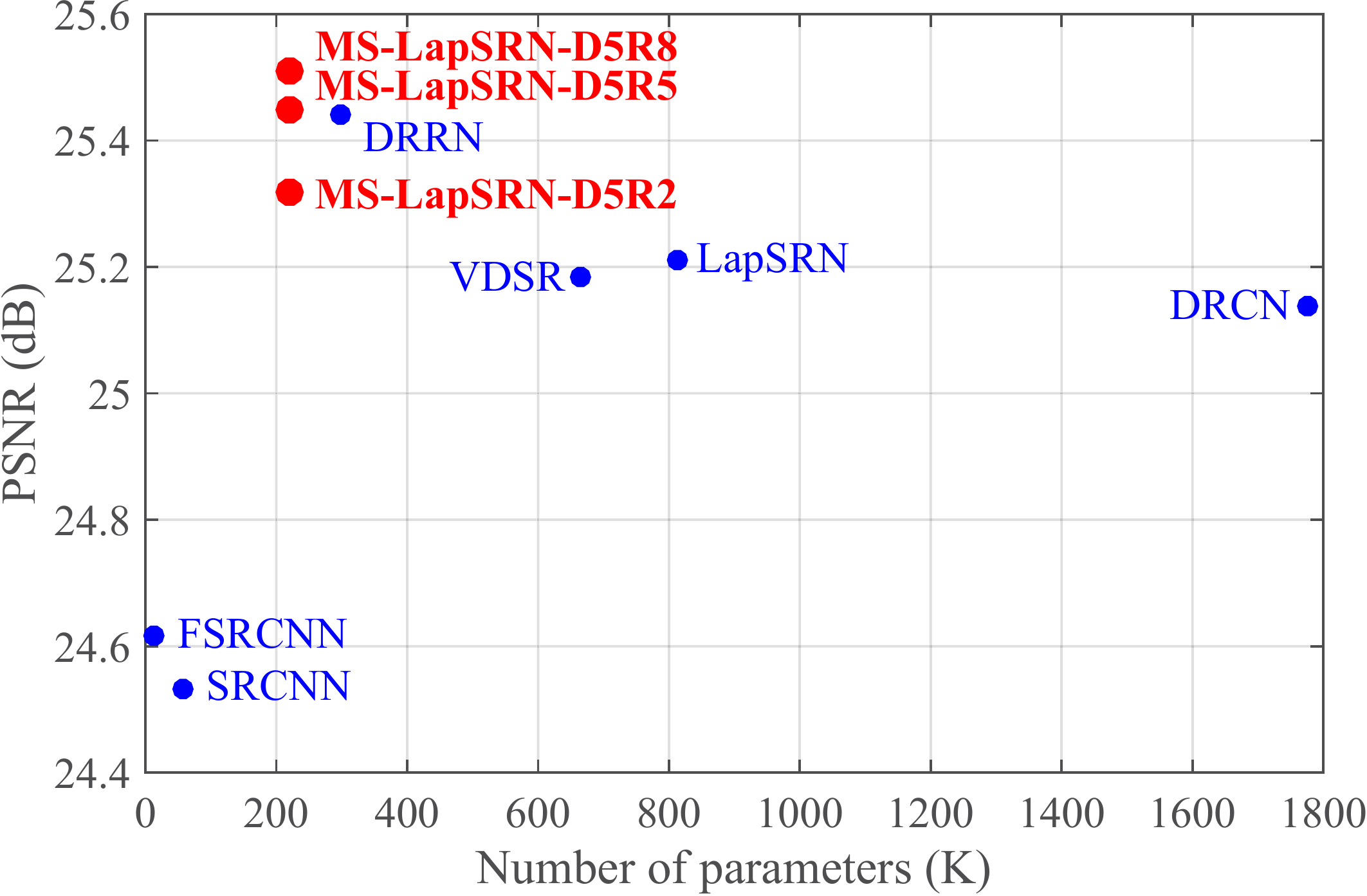}
	\caption{
		\textbf{Number of network parameters versus performance}.
		The results are evaluated on the \textsc{Urban100} dataset for $4\times$ SR. 
		The proposed MS-LapSRN strides a balance between reconstruction accuracy and execution time.
	}
	\label{fig:psnr_parameters}
\end{figure}

We show the reconstruction performance versus the number of network parameters of CNN-based SR methods in~\figref{psnr_parameters}.
By sharing parameters and using recursive layers, our MS-LapSRN has parameters about $73\%$ less than the LapSRN~\cite{LapSRN}, $66\%$ less than the VDSR~\cite{VDSR}, $87\%$ less than the DRCN~\cite{DRCN}, and $25\%$ less than the DRRN~\cite{DRRN}.
While our model has a smaller footprint, we achieve the state-of-the-art performance among these CNN-based methods.
Comparing to the SRCNN~\cite{SRCNN} and FSRCNN~\cite{FSRCNN}, our MS-LapSRN-D5R8 has about 0.9 to 1 dB improvement on the challenging \textsc{Urban100} dataset for $4\times$ SR.

\begin{figure}
	\footnotesize
	\centering
	\hspace{-0.2cm}
	\begin{adjustbox}{valign=t}
		\begin{tabular}{c}
		\includegraphics[width=0.385\columnwidth]{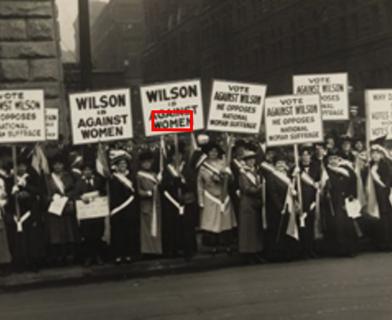}
		\\
		Input LR
		\end{tabular}
	\end{adjustbox}
	\hspace{-0.4cm}
	\begin{adjustbox}{valign=t}
		\begin{tabular}{cc}
		\includegraphics[width=0.265\columnwidth]{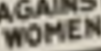} \hspace{-0.3cm} &
		\includegraphics[width=0.265\columnwidth]{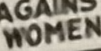}
		\\
		Bicubic \hspace{-0.3cm} &
		FSRCNN~\cite{FSRCNN}
		\\
		\includegraphics[width=0.265\columnwidth]{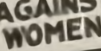} \hspace{-0.3cm} &
		\includegraphics[width=0.265\columnwidth]{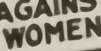} 
		\\
		\hspace{0.2cm} VDSR~\cite{VDSR}&
		MS-LapSRN (ours)
		\end{tabular}
	\end{adjustbox}
	\\
	\vspace{1mm}
	\hspace{-0.2cm}
	\begin{adjustbox}{valign=t}
		\begin{tabular}{c}
		\includegraphics[width=0.39\columnwidth]{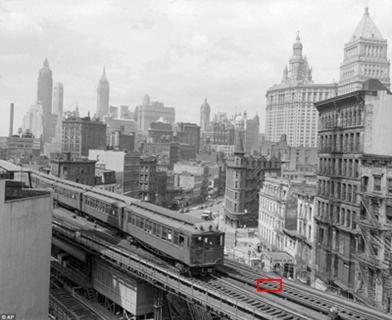}
		\\
		Input LR
		\end{tabular}
	\end{adjustbox}
	\hspace{-0.4cm}
	\begin{adjustbox}{valign=t}
		\begin{tabular}{cc}
		\includegraphics[width=0.265\columnwidth]{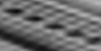} \hspace{-0.3cm} &
		\includegraphics[width=0.265\columnwidth]{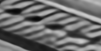}
		\\
		Bicubic \hspace{-0.3cm} &
		DRCN~\cite{DRCN}
		\\
		\includegraphics[width=0.265\columnwidth]{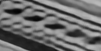} \hspace{-0.3cm} &
		\includegraphics[width=0.265\columnwidth]{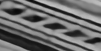} 
		\\
		\hspace{0.2cm} LapSRN~\cite{LapSRN}&
		MS-LapSRN (ours)
		\end{tabular}
	\end{adjustbox}
	\vspace{-2mm}
	\caption{
		\textbf{Comparison of real-world photos for $4\times$ SR.} 
		The ground truth HR images and the blur kernels are not available in these cases.
		On the top image, our method super-resolves the letter ``W'' accurately while VDSR incorrectly connects the stroke with the letter ``O''.
		On the bottom image, our method reconstructs the rails without the artifacts.
	}
	\label{fig:historical}
	\vspace{-2mm}
\end{figure}

\subsection{Super-resolving real-world photos}
We demonstrate an application of super-resolving historical photographs with JPEG compression artifacts.
In these cases, neither the ground-truth images nor the downsampling kernels are available. 
As shown in~\figref{historical}, our method can reconstruct sharper and more accurate images than the state-of-the-art approaches.

\vspace{-2mm}
\subsection{Comparison to LAPGAN}
As described in~\secref{related_lapalcian}, the target applications of the LAPGAN~\cite{LAPGAN} and LapSRN are different.
Therefore, we focus on comparing the \emph{network architectures} for image super-resolution.
We train the LAPGAN and LapSRN for $4\times$ and $8\times$ SR with the same training data and settings.
We use 5 convolutional layers at each level and optimize both networks with the Charbonnier loss function.
We note that in~\cite{LAPGAN} the sub-networks are independently trained.
For fair comparisons, we jointly train the entire network for both LAPGAN and LapSRN.
We present quantitative comparisons and runtime on the \textsc{Set14} and \textsc{BSDS100} datasets in~\tabref{LAPGAN}.
Under the same training setting, our method achieves more accurate reconstruction and faster execution speed than that of the LAPGAN.

\begin{table}
	\centering
	\caption{
		\textbf{Quantitative comparisons between the generative network of the LAPGAN~\cite{LAPGAN} and our LapSRN.} 
		Our LapSRN achieves better reconstruction quality and faster processing speed than the LAPGAN.
	}
	\vspace{-3mm}
	\begin{tabular}{c|c|cc|cc}
		\toprule
		\multirow{2}{*}{Method} &
		\multirow{2}{*}{Scale} &
		\multicolumn{2}{c|}{\textsc{Set14}} &
		\multicolumn{2}{c}{\textsc{BSDS100}}
		\\ 
		& & PSNR & Seconds & PSNR & Seconds
		\\ 
		\midrule
		LAPGAN & 4 & 
		27.89 & 0.0446 & 
		27.09 & 0.0135 \\
		LapSRN & 4 & 
		28.04 & 0.0395 & 
		27.22 & 0.0078 \\
		\midrule
		LAPGAN & 8 & 
		24.30 & 0.0518 & 
		24.46 & 0.0110 \\
		LapSRN & 8 & 
		24.42 & 0.0427 & 
		24.53 & 0.0107 \\
		\bottomrule
	\end{tabular}
	\label{tab:LAPGAN}
	\vspace{-2mm}
\end{table}

\begin{figure}
	\centering
	\footnotesize
	\vspace{-2mm}
	\begin{tabular}{ccc}
		\hspace{-2mm}
		\includegraphics[width=0.31\linewidth]{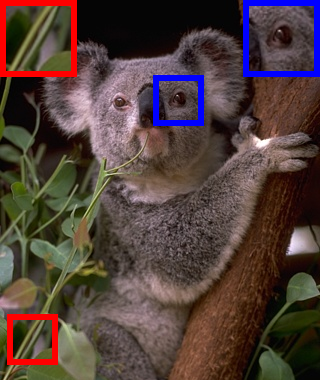} 
		& \hspace{-3mm}
		\includegraphics[width=0.31\linewidth]{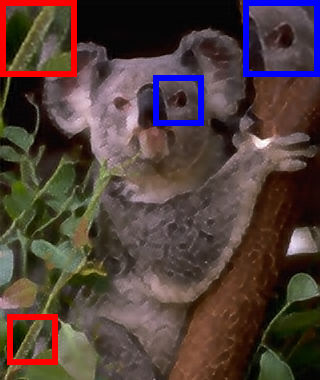} 
		& \hspace{-3mm}
		\includegraphics[width=0.31\linewidth]{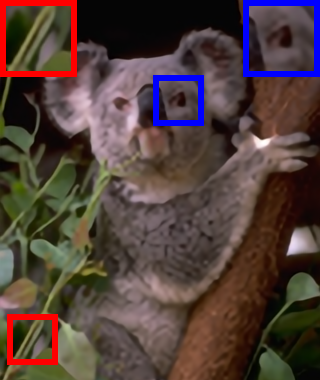} 
		\\
		\hspace{-2mm}
		(a) Ground Truth HR 
		& \hspace{-3mm} 
		(b) LapSRN + adv. 
		& \hspace{-3mm} 
		(c) LapSRN 
	\end{tabular}
	\vspace{-0.3cm}
	\caption{
		\textbf{Visual comparison for adversarial training}. 
		We compare the results trained with and without the adversarial training on $4\times$ SR.
	}
	\label{fig:visual_adv}
	\vspace{-2mm}
\end{figure}

\subsection{Adversarial training}
\label{sec:adversarial}
We demonstrate that our LapSRN can be extended to incorporate the adversarial training~\cite{GAN}.
We treat our LapSRN as a generative network and build a discriminative network using the discriminator of the DCGAN~\cite{DCGAN}.
The discriminative network consists of four convolutional layers with a stride of 2, two fully connected layers and one sigmoid layer to generate a scalar probability for distinguishing between real images and generated images from the generative network.
We find that it is difficult to obtain accurate SR images by solely minimizing the cross-entropy loss.
Therefore, we include the pixel-wise reconstruction loss (i.e., Charbonnier loss) to enforce the similarity between the input LR images and the corresponding ground truth HR images.

We show a visual result in~\figref{visual_adv} for $4\times$ SR.
The network with the adversarial training generates more plausible details on regions of irregular structures, e.g., grass, and feathers.
However, the predicted results may not be faithfully reconstructed with respect to the ground truth high-resolution images. 
As a result, the accuracy is not as good as the model trained with the Charbonnier loss.

\subsection{Limitations}
\begin{figure}
	\footnotesize
	\centering
	\begin{tabular}{cc}
		\hspace{-0.45cm}
		\begin{adjustbox}{valign=t}
			\begin{tabular}{c}
				\includegraphics[width=0.3\columnwidth]{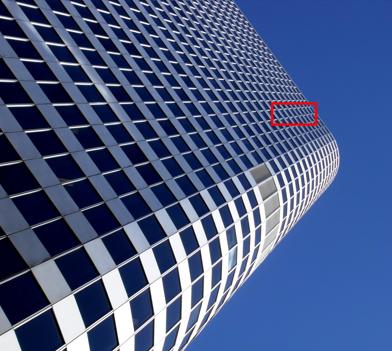}
				\\
				Ground-truth HR
			\end{tabular}
		\end{adjustbox}
		\hspace{-0.5cm}
		\begin{adjustbox}{valign=t}
			\begin{tabular}{ccc}
				\includegraphics[width=0.22\columnwidth]{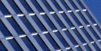} \hspace{-0.4cm} &
				\includegraphics[width=0.22\columnwidth]{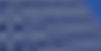} \hspace{-0.4cm} &
				\includegraphics[width=0.22\columnwidth]{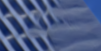}
				\\
				HR \hspace{-0.4cm} &
				Bicubic \hspace{-0.4cm} &
				SelfExSR~\cite{SelfExSR}
				\\
				\includegraphics[width=0.22\columnwidth]{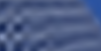} \hspace{-0.4cm} &
				\includegraphics[width=0.22\columnwidth]{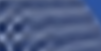} \hspace{-0.4cm} &
				\includegraphics[width=0.22\columnwidth]{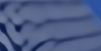} 
				\\
				VDSR~\cite{VDSR} \hspace{-0.4cm} &
				DRRN~\cite{DRRN} \hspace{-0.4cm} &
				MS-LapSRN
			\end{tabular}
		\end{adjustbox}
	\end{tabular}
	\vspace{-2mm}
	\caption{
		\textbf{Limitation.} A failure case for $8\times$ SR. Our method is not able to hallucinate details if the LR input image does not consist of sufficient amount of structure.}
	\label{fig:failure}
\end{figure}

While our model is capable of generating clean and sharp HR images for large upsampling scales, e.g., $8\times$, it does not ``hallucinate'' fine details.
As shown in~\figref{failure}, the top of the building is significantly blurred in the $8\times$ downscaled LR image.
All SR algorithms fail to recover the fine structure except the SelfExSR~\cite{SelfExSR} method which explicitly detects the 3D scene geometry and uses self-similarity to hallucinate the regular structure.
This is a common limitation shared by parametric SR methods~\cite{A+,SRCNN,FSRCNN,VDSR,DRCN,DRRN}.

\vspace{-3mm}
\section{Conclusions}
In this work, we propose a deep convolutional network within a Laplacian pyramid framework for fast and accurate image super-resolution.
Our model progressively predicts high-frequency residuals in a coarse-to-fine manner with deeply supervision from the robust Charbonnier loss functions.
%
%
By sharing parameters \emph{across} as well as \emph{within} pyramid levels, we use $73\%$ fewer parameters than our preliminary method~\cite{LapSRN} and achieve improved performance.
We incorporate local skip connections in our network for training deeper models.
Furthermore, we adopt the multi-scale training strategy to train a \emph{single} model for handling \emph{multiple} upsampling scales.
We present a comprehensive evaluation on various design choices and believe that the thorough analysis benefits our community.
%
%
We have shown the promise of the proposed LapSRN in the context of image super-resolution.
Yet, our network design is general and can potentially be applied to other image transformation and synthesis problems.

\section{Appendix}

\subsection{Multi-scale Training}
\begin{figure}
	\footnotesize
	\centering
	\hspace{-0.2cm} 
	\begin{adjustbox}{valign=t}
		\begin{tabular}{c}
			\includegraphics[width=0.345\columnwidth]{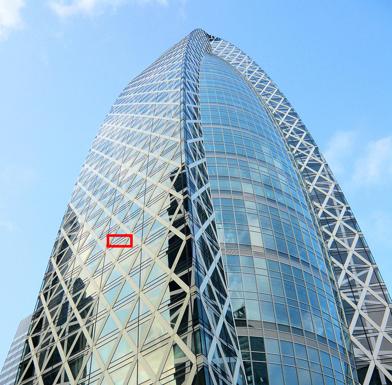}
			\\
			Ground-truth HR
		\end{tabular}
	\end{adjustbox}
	\hspace{-0.4cm}
	\begin{adjustbox}{valign=t}
		\begin{tabular}{cc}
			\includegraphics[width=0.29\columnwidth]{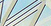} \hspace{-0.3cm} &
			\includegraphics[width=0.29\columnwidth]{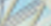}
			\\
			HR \hspace{-0.3cm} &
			Bicubic
			\\
			\includegraphics[width=0.29\columnwidth]{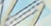} \hspace{-0.3cm} &
			\includegraphics[width=0.29\columnwidth]{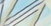} 
			\\
			Train on $4\times$ \hspace{-0.3cm} &
			Train on $2\times, 4\times, 8\times$
		\end{tabular}
	\end{adjustbox}
	\\
	\vspace{1mm}
	(a) $3\times$ SR
	\vspace{1mm}
	\\
	\hspace{-0.2cm}
	\begin{adjustbox}{valign=t}
		\begin{tabular}{c}
			\includegraphics[width=0.345\columnwidth]{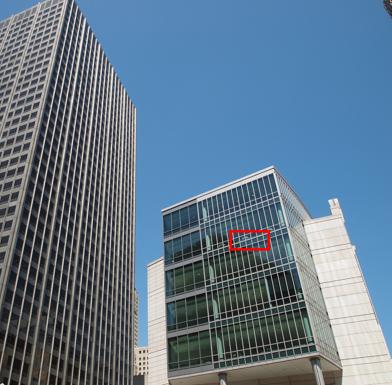}
			\\
			Ground-truth HR
		\end{tabular}
	\end{adjustbox}
	\hspace{-0.4cm}
	\begin{adjustbox}{valign=t}
		\begin{tabular}{cc}
			\includegraphics[width=0.29\columnwidth]{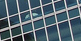} \hspace{-0.3cm} &
			\includegraphics[width=0.29\columnwidth]{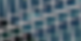}
			\\
			HR \hspace{-0.3cm} &
			Bicubic
			\\
			\includegraphics[width=0.29\columnwidth]{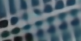} \hspace{-0.3cm} &
			\includegraphics[width=0.29\columnwidth]{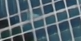} 
			\\
			Train on $4\times$ \hspace{-0.3cm} &
			Train on $2\times, 4\times, 8\times$
		\end{tabular}
	\end{adjustbox}
	\\
	\vspace{1mm}
	(b) $4\times$ SR
	\vspace{1mm}
	\\
	\hspace{-0.2cm} 
	\begin{adjustbox}{valign=t}
		\begin{tabular}{c}
			\includegraphics[width=0.345\columnwidth]{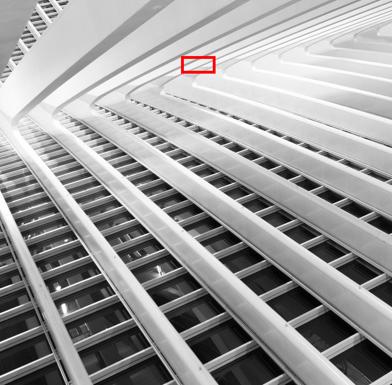}
			\\
			Ground-truth HR
		\end{tabular}
	\end{adjustbox}
	\hspace{-0.4cm}
	\begin{adjustbox}{valign=t}
		\begin{tabular}{cc}
			\includegraphics[width=0.29\columnwidth]{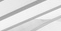} \hspace{-0.3cm} &
			\includegraphics[width=0.29\columnwidth]{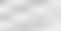}
			\\
			HR \hspace{-0.3cm} &
			Bicubic
			\\
			\includegraphics[width=0.29\columnwidth]{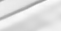} \hspace{-0.3cm} &
			\includegraphics[width=0.29\columnwidth]{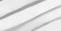} 
			\\
			Train on $8\times$ \hspace{-0.3cm} &
			Train on $2\times, 4\times, 8\times$
		\end{tabular}
	\end{adjustbox}
	\\
	\vspace{1mm}
	(c) $8\times$ SR
	\vspace{1mm}
	\\
	\vspace{-3mm}
	\caption{
		\textbf{Visual comparison of multi-scale training.} 
		The model using the multi-scale training restores more details. 
	}
	\label{fig:multiscale}
\end{figure}

We train our $\text{LapSRN}_{\textbf{SS}}$-D5R8 model with the following combinations of upsampling scales: $\{2\times\}$, $\{4\times\}$, $\{8\times\}$, $\{2\times, 4\times\}$, $\{2\times, 8\times\}$, $\{4\times, 8\times\}$ and $\{2\times, 4\times, 8\times\}$.
We show quantitative evaluation on the \textsc{Set14}~\cite{Zeyde-2010}, \textsc{BSDS100}~\cite{BSDS} and \textsc{Urban100}~\cite{SelfExSR} datasets in~\tabref{multiscale} and visual comparisons in~\figref{multiscale}..
In general, the multi-scale models perform better than single-scale models.
Note that our models do not use any $3\times$ SR samples for training.
Although our $4\times$ model is able to generate decent results for $3\times$ SR, the multi-scale models, especially the models trained on $\{2\times, 4\times\}$ and $\{2\times, 4\times, 8\times\}$ SR, further improve the performance by 0.14 and 0.17 dB on the \textsc{Urban100} dataset, respectively.

%
%

\begin{table*}
	\centering
	\footnotesize
	\caption{
		\textbf{Quantitative evaluation of multi-scale training.}
		We train the proposed model with combinations of $2\times, 4\times$ and $8\times$ SR samples and evaluate on the \textsc{Set14}, \textsc{BSDS100} and \textsc{Urban100} 
		datasets for $2\times, 3\times, 4\times$ and $8\times$ SR.
		The model trained with $2\times, 4\times$ and $8\times$ SR samples together achieves better performance on all upsampling scales and can also generalize to unseen $3\times$ SR examples.
	}
	\vspace{-3mm}
	\begin{tabular}{l|cccc|cccc|cccc}
		\toprule
		\multirow{2}{*}{Train $\backslash$ Test} &
		\multicolumn{4}{c|}{\textsc{Set14}} &
		\multicolumn{4}{c|}{\textsc{BSDS100}} &
		\multicolumn{4}{c}{\textsc{Urban100}} \\
		& 
		$2\times$ & $3\times$ & $4\times$ & $8\times$ &
		$2\times$ & $3\times$ & $4\times$ & $8\times$ &
		$2\times$ & $3\times$ & $4\times$ & $8\times$
		\\
		\midrule
		$2\times$
		&
		33.24 & 27.31 & 25.21 & 21.81 & 32.01 & 26.93 & 25.13 & 22.25 & 31.01 & 24.93 & 22.80 & 19.89
		\\
		$4\times$
		&
		32.96 & 29.90 & 28.21 & 23.97 & 31.80 & \second{28.89} & 27.39 & 24.26 & 30.53 & 27.30 & 25.38 & 21.61
		\\
		$8\times$
		&
		32.36 & 29.89 & 28.14 & \second{24.52} & 31.28 & 28.84 & 27.34 & \second{24.63} & 29.16 & 27.15 & 25.23 & 21.95
		\\
		$2\times, 4\times$
		&
		\second{33.25} & \second{29.96} & \first{28.27} & 24.15 & \second{32.04} & \first{28.93} & \second{27.42} & 24.32 & \first{31.17} & \second{27.44} & \second{25.49} & 21.81
		\\
		$2\times, 8\times$
		&
		33.22 & 29.93 & 28.17 & \second{24.52} & 32.01 & \second{28.89} & 27.38 & \second{24.63} & 31.05 & 27.32 & 25.38 & 22.00
		\\
		$4\times, 8\times$
		&
		32.88 & 29.90 & 28.20 & 24.48 & 31.76 & \second{28.89} & 27.40 & 24.62 & 30.42 & 27.37 & 25.43 & \second{22.01}
		\\
		$2\times, 4\times, 8\times$
		&
		\first{33.28} & \first{29.97} & \second{28.26} & \first{24.57} & \first{32.05} & \first{28.93} & \first{27.43} & \first{24.65} & \second{31.15} & \first{27.47} & \first{25.51} & \first{22.06}
		\\
		\bottomrule
	\end{tabular}
	\label{tab:multiscale}
	\vspace{-3mm}
\end{table*}

\subsection{Human Subject Study}
To measure the human perceptual preferences on super-resolved images, we conduct a human subject study to evaluate several state-of-the-art SR algorithms.
A straightforward strategy is to ask users to give an \emph{absolute} score (e.g., 1 to 5) or provide ranking on all SR results for each test image.
One can then compute the average score for each method.
However, such scores might not be sufficiently reliable when there are a large number of images to be compared.
Furthermore, as super-resolved images from different algorithms often have subtle differences, it is difficult for users to make comparisons on multiple images simultaneously.

\begin{figure}
	\centering		
	\includegraphics[width=0.85\columnwidth]{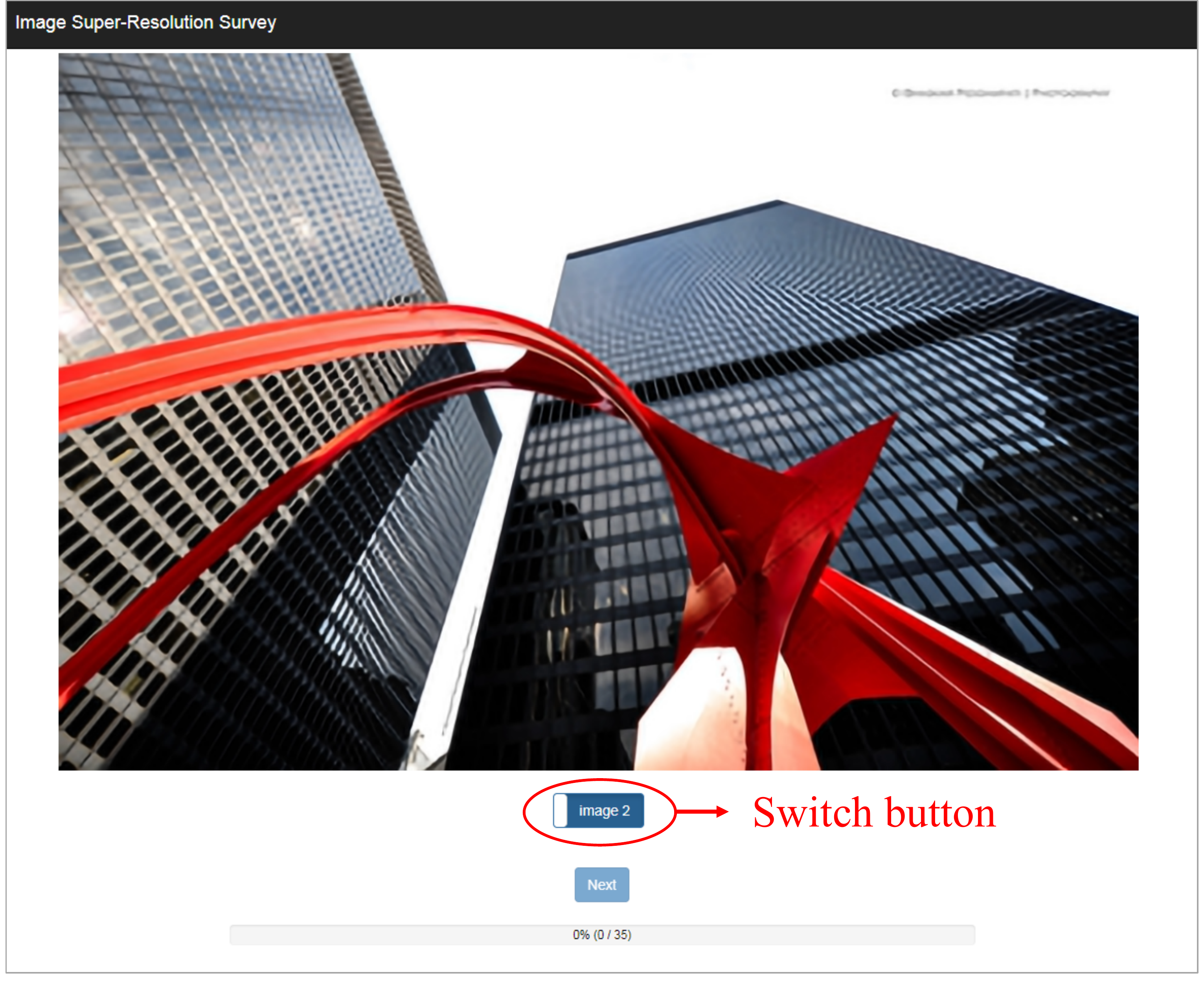}
	\caption{
		\textbf{Interface for our human subject study.}
		Human subjects can switch back and forth between two given images (results from two different super-resolution algorithms) to see the differences.
	}
	\label{fig:interface}
	\vspace{-2mm}
\end{figure}

In light of this, we choose to conduct the \emph{paired} comparison for our subject study.
Paired comparison is also adopted to evaluate the perceptual quality of image retargeting~\cite{Rubinstein-TOG-2010} and image deblurring~\cite{Lai-CVPR-2016}.
For each test, each user is asked to select the preferred one from a pair of images.
We design a web-based interface (\figref{interface}) for users to switch back and forth between two given images.
Users can easily see the differences between the two images and make selections.
Through the study, we obtain the \emph{relative} scores between every pair of evaluated methods.
We conduct such paired comparison for FSRCNN~\cite{FSRCNN}, VDSR~\cite{VDSR}, LapSRN~\cite{LapSRN}, DRRN~\cite{DRRN} and the proposed MS-LapSRN on the BSDS100~\cite{BSDS} and Urban100~\cite{SelfExSR} datasets for $4\times$ SR.

We ask each participant to compare 30 pairs of images.
To detect casual or careless users, we include 5 pairs of images as the sanity check.
In these image pairs, we show the ground truth HR and the bicubic upsampled images.
The users must select the ground truth HR image to pass the sanity check.
We discard the results by a subject if the subject fails the sanity check more than twice.
Finally, we collect the results from 71 participants.

\begin{figure}
	\centering
	\includegraphics[width=0.85\columnwidth]{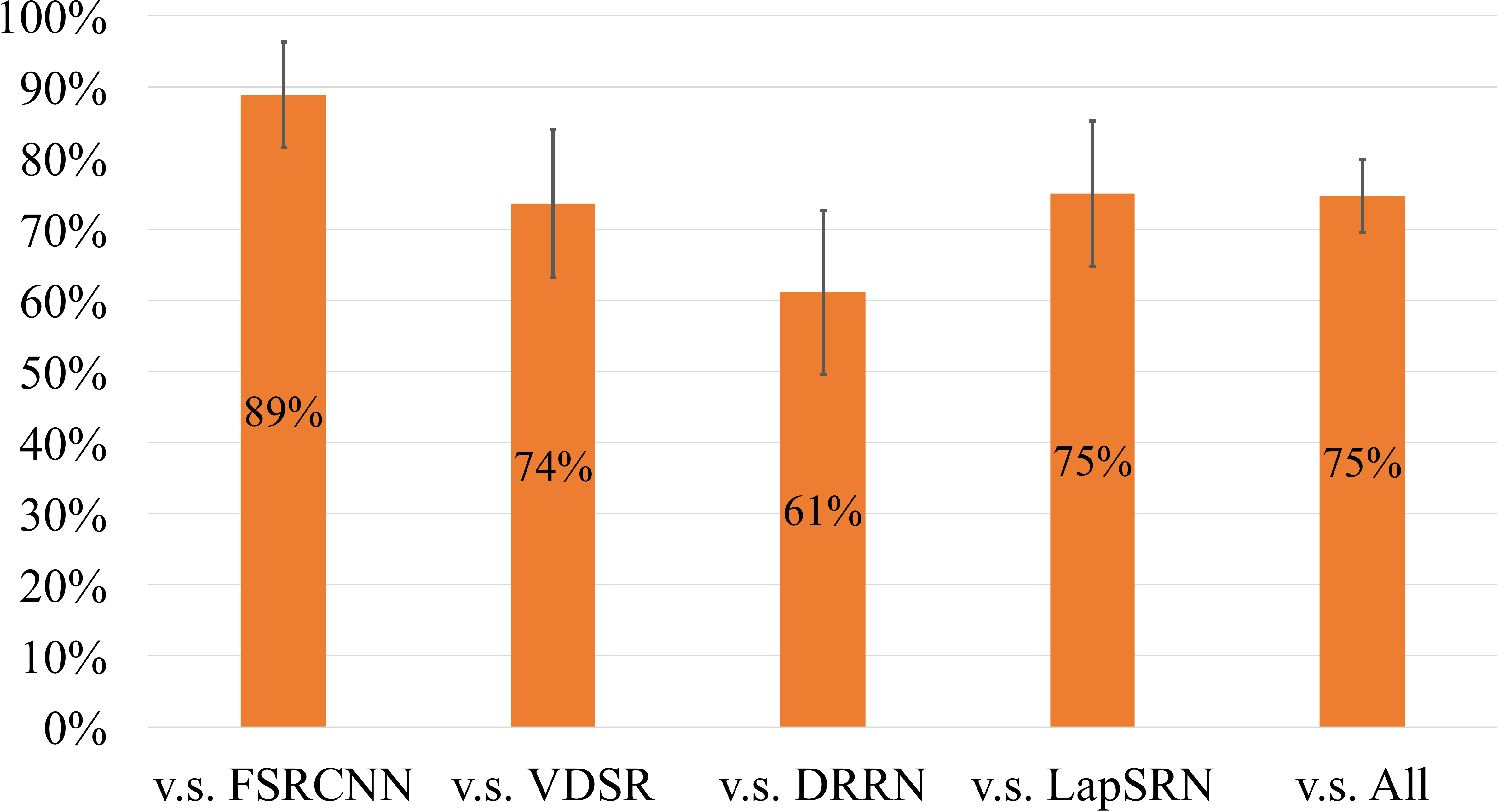}
	\\
	(a) Results on \textsc{BSDS100}
	\\	
	\includegraphics[width=0.85\columnwidth]{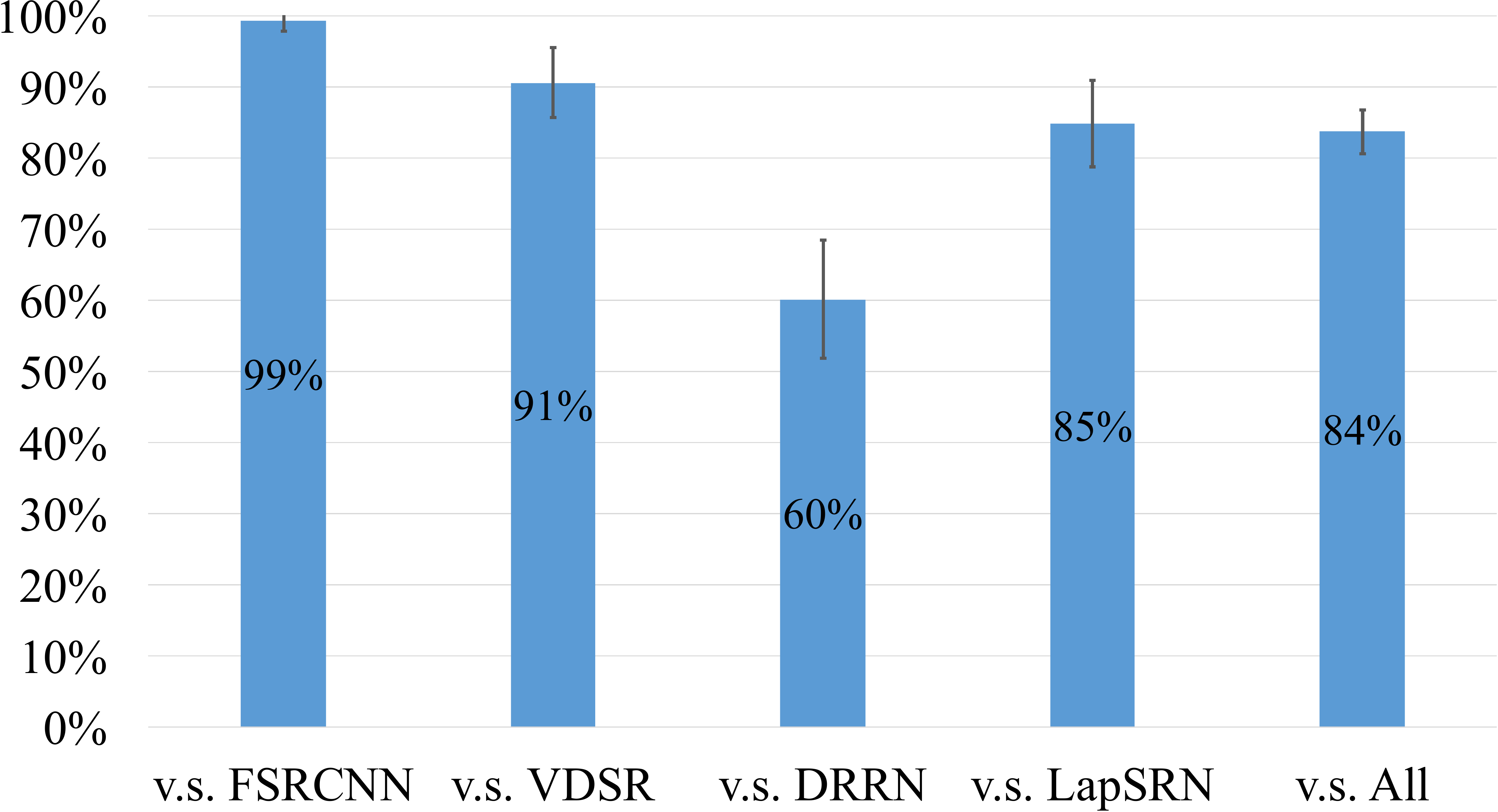}
	\\
	(b) Results on \textsc{Urban100}
	\caption{
		\textbf{Analysis on human subject study.}
		Our MS-LapSRN is preferred by $75\%$ and $80\%$ of users on the \textsc{BSDS100} and \textsc{Urban100} datasets, respectively.
		The error bars show the $95\%$ confidence interval. 
	}
	\label{fig:userstudy}
\end{figure}

To obtain a global score for each method, we fit the results of paired comparisons to the Bradley-Terry (BT) model~\cite{BT-model}.
We refer readers to~\cite{Rubinstein-TOG-2010,Lai-CVPR-2016} for details of the BT model.
We normalize the BT scores to zero means and show the results in~\tabref{BT}.
The proposed MS-LapSRN performs favorably against other approaches on both the BSDS100 and Urban100 datasets.

\begin{table}
	\centering
	\caption{
	\textbf{BT scores of SR algorithms in human subject study.} 
	Our MS-LapSRN performs favorably against other compared methods.
	}
	\vspace{-3mm}
	\begin{tabular}{r|cc}
		\toprule
		Method &
		\textsc{BSDS100} &
		\textsc{Urban100}
		\\ 
		\midrule
		FSRCNN~\cite{FSRCNN} 
		& -1.1291 & -1.8005 \\
		VDSR~\cite{VDSR} 
		& 0.0357 & 0.0981 \\
		LapSRN~\cite{LapSRN} 
		& 0.1910 & 0.2415\\ 
		DRRN~\cite{DRRN} 
		& 0.3721 & 0.6521 \\ 
		MS-LapSRN 
		& 0.5304 & 0.8087 \\ 
		\bottomrule
	\end{tabular}
	\label{tab:BT}
	\vspace{-3mm}
\end{table}

We further analyze the comparisons between our MS-LapSRN and other methods.
We compute the percentage that users choose our MS-LapSRN over FSRCNN, VDSR, DRRN, and LapSRN and plot the results in~\figref{userstudy} (a) and (b).
On average, our MS-LapSRN is preferred by $75\%$ of users on the \textsc{BSDS100} dataset and $84\%$ of users on the \textsc{Urban100} dataset.
When comparing with DRRN, our MS-LapSRN obtains around $60\%$ of votes as the performance is close to each other.
The human subject study also shows that the results of our MS-LapSRN have higher perceptual quality than existing methods.

\begin{table*}
	\footnotesize
	\centering
	\caption{
		\textbf{Quantitative evaluation on the \textsc{DIV2K} dataset}. 
		\red{\textbf{Red}} and \blue{\underline{blue}} indicate the best and the second best performance, respectively.
	}
	\vspace{-3mm}
	\label{tab:div2k} 
	\begin{tabular}{rcccc}
			\toprule
			\multirow{2}{*}{Algorithm}
			& 
			$2\times$ & $3\times$ & $4\times$ & $8\times$
			\\
			&
			PSNR / SSIM / IFC & 
			PSNR / SSIM / IFC & 
			PSNR / SSIM / IFC & 
			PSNR / SSIM / IFC \\
			\midrule
			Bicubic & 
			32.45 / 0.904 / 6.348 &
			29.66 / 0.831 / 3.645 &
			28.11 / 0.775 / 2.352 &
			25.18 / 0.247 / 0.837
			\\
			A+~\cite{A+} & 
			34.56 / 0.933 / 8.314 &
			31.09 / 0.865 / 4.618 &
			29.28 / 0.809 / 3.004 &
			25.93 / 0.686 / 1.024
			\\
			RFL~\cite{RFL} & 
			34.48 / 0.932 / 8.299 &
			31.01 / 0.863 / 4.601 &
			29.19 / 0.806 / 2.956 &
			25.82 / 0.679 / 0.949
			\\
			SelfExSR~\cite{SelfExSR} & 
			34.54 / 0.933 / 7.880 &
			31.16 / 0.867 / 4.426 &
			29.35 / 0.812 / 2.891 &
			25.94 / 0.687 / 0.981
			\\
			SRCNN~\cite{SRCNN} & 
			34.59 / 0.932 / 7.143 &
			31.11 / 0.864 / 3.992 &
			29.33 / 0.809 / 2.639 &
			26.05 / 0.691 / 0.974
			\\
			FSRCNN~\cite{FSRCNN} & 
			34.74 / 0.934 / 8.004 &
			31.25 / 0.868 / 4.442 &
			29.36 / 0.811 / 2.792 &
			25.77 / 0.682 / 0.872
			\\
			SCN~\cite{SCN} & 
			34.98 / 0.937 / 7.981 &
			31.42 / 0.870 / 4.632 &
			29.47 / 0.813 / 2.728 &
			26.01 / 0.687 / 0.938
			\\
			VDSR~\cite{VDSR} & 
			35.43 / \second{0.941} / 8.390 &
			31.76 / \second{0.878} / 4.766 &
			29.82 / 0.824 / 3.004 &
			26.23 / 0.699 / 1.063
			\\
			DRCN~\cite{DRCN} & 
			35.45 / 0.940 / 8.618 &
			31.79 / 0.877 / \first{4.854} &
			29.83 / 0.823 / 3.082 &
			26.24 / 0.698 / 1.066
			\\
			LapSRN~\cite{LapSRN} & 
			35.31 / 0.940 / 8.587 &
			31.22 / 0.861 / 4.534 &
			29.88 / 0.825 / 3.131 &
			26.11 / 0.697 / 1.114
			\\
			DRRN~\cite{DRRN} & 
			\first{35.63} / \second{0.941} / 8.246 &
			\first{31.96} / \first{0.880} / \second{4.804} &
			29.98 / 0.827 / 3.156 &
			26.37 / 0.704 / 1.101
			\\
			MS-LapSRN-D5R2 (ours) & 
			35.45 / \second{0.941} / 8.603 &
			31.72 / 0.874 / 4.599 &
			29.92 / 0.826 / 3.127 &
			\second{26.43} / 0.708 / 1.166
			\\
			MS-LapSRN-D5R5 (ours) & 
			\second{35.58} / \second{0.941} / \second{8.789} &
			31.82 / 0.875 / 4.719 &
			\second{30.01} / \second{0.828} / \second{3.204} &
			\first{26.53} / \second{0.710} / \second{1.200}
			\\
			MS-LapSRN-D5R8 (ours) & 
			\first{35.63} / \first{0.942} / \first{8.799} &
			\second{31.86} / 0.876 / 4.766 &
			\first{30.04} / \first{0.829} / \first{3.233} &
			\first{26.53} / \first{0.711} / \first{1.209}
			\\
			\bottomrule 
		\end{tabular}
	\vspace{-1mm}
\end{table*}

\subsection{Results on DIV2K}
We evaluate our MS-LapSRN on the DIV2K dataset~\cite{DIV2K} in~\tabref{div2k}.
The proposed method performs favorably against existing approaches on $2\times$, $4\times$, and $8\times$ SR. 



\bibliographystyle{IEEEtran}
\bibliography{super_resolution}
%



%

\vspace{-1.2cm}

\begin{IEEEbiography}[{\includegraphics[width=0.9in,clip,keepaspectratio]{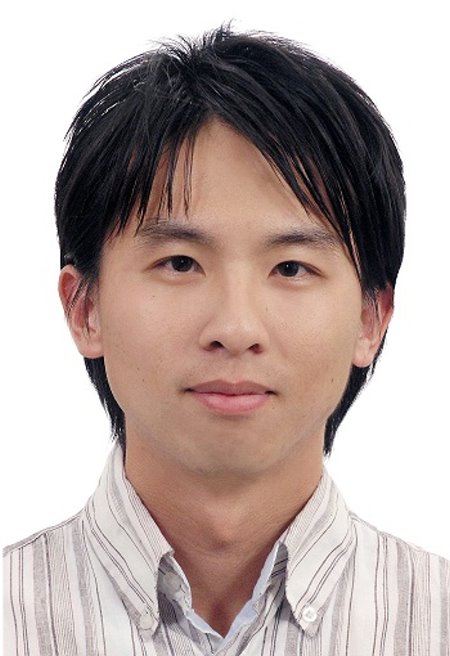}}]
{Wei-Sheng Lai} is a Ph.D. candidate of Electrical Engineering and Computer Science at the University of California, Merced, CA, USA. He received the B.S. and M.S. degree in Electrical Engineering from the National Taiwan University, Taipei, Taiwan, in 2012 and 2014, respectively. His research interests include computer vision, computational photography, and deep learning.
\end{IEEEbiography}

\vspace{-1.0cm}

\begin{IEEEbiography}[{\includegraphics[width=0.9in,clip,keepaspectratio]{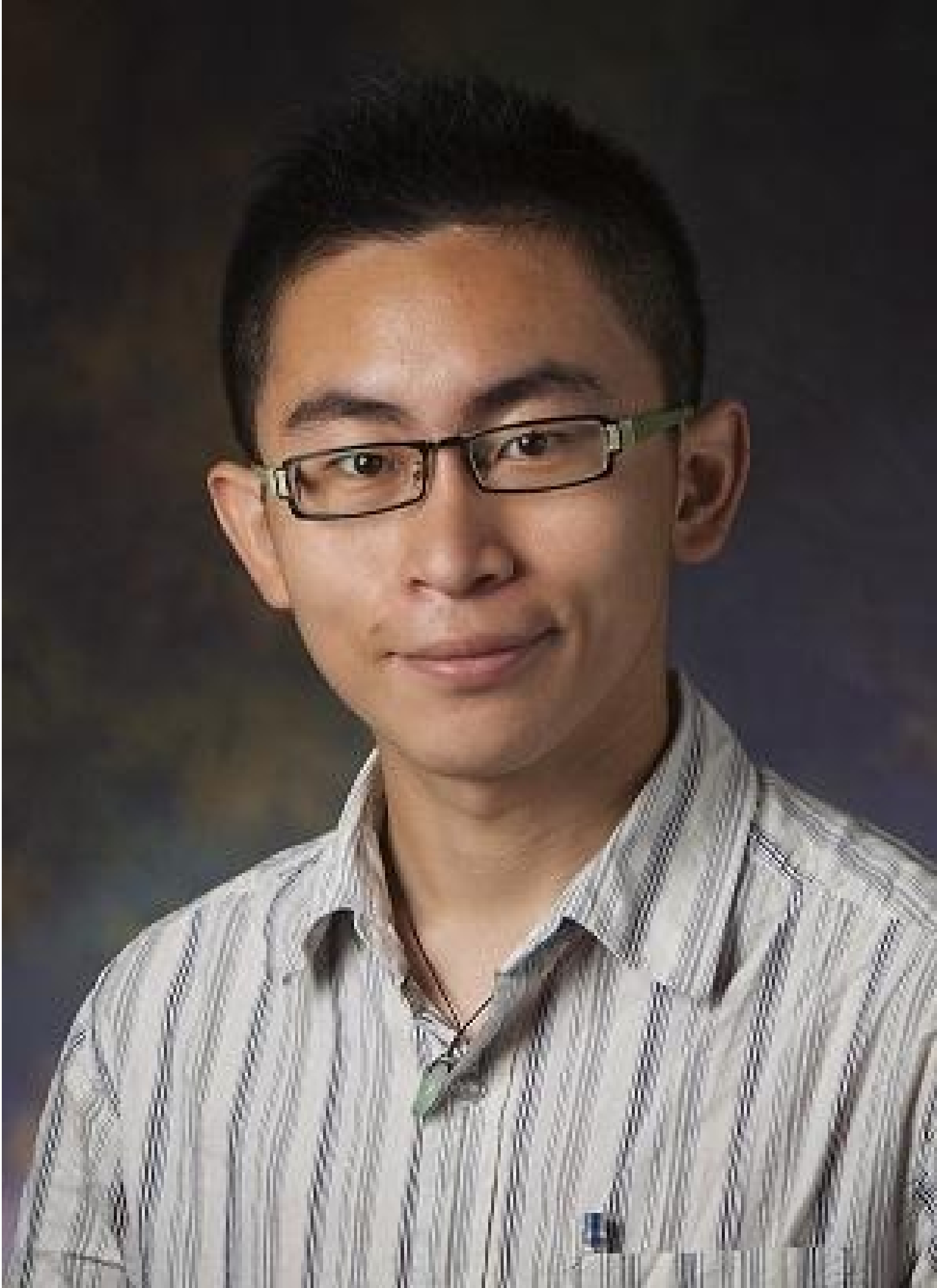}}]
{Jia-Bin Huang} is an assistant professor in the Bradley Department of Electrical and Computer Engineering at Virginia Tech. He received the B.S. degree in Electronics Engineering from National Chiao-Tung University, Hsinchu, Taiwan and his Ph.D. degree in the Department of Electrical and Computer Engineering at University of Illinois, Urbana-Champaign in 2016. He is a member of the IEEE.
\end{IEEEbiography}

\vspace{-1.0cm}

\begin{IEEEbiography}[{\includegraphics[width=0.9in,clip,keepaspectratio]{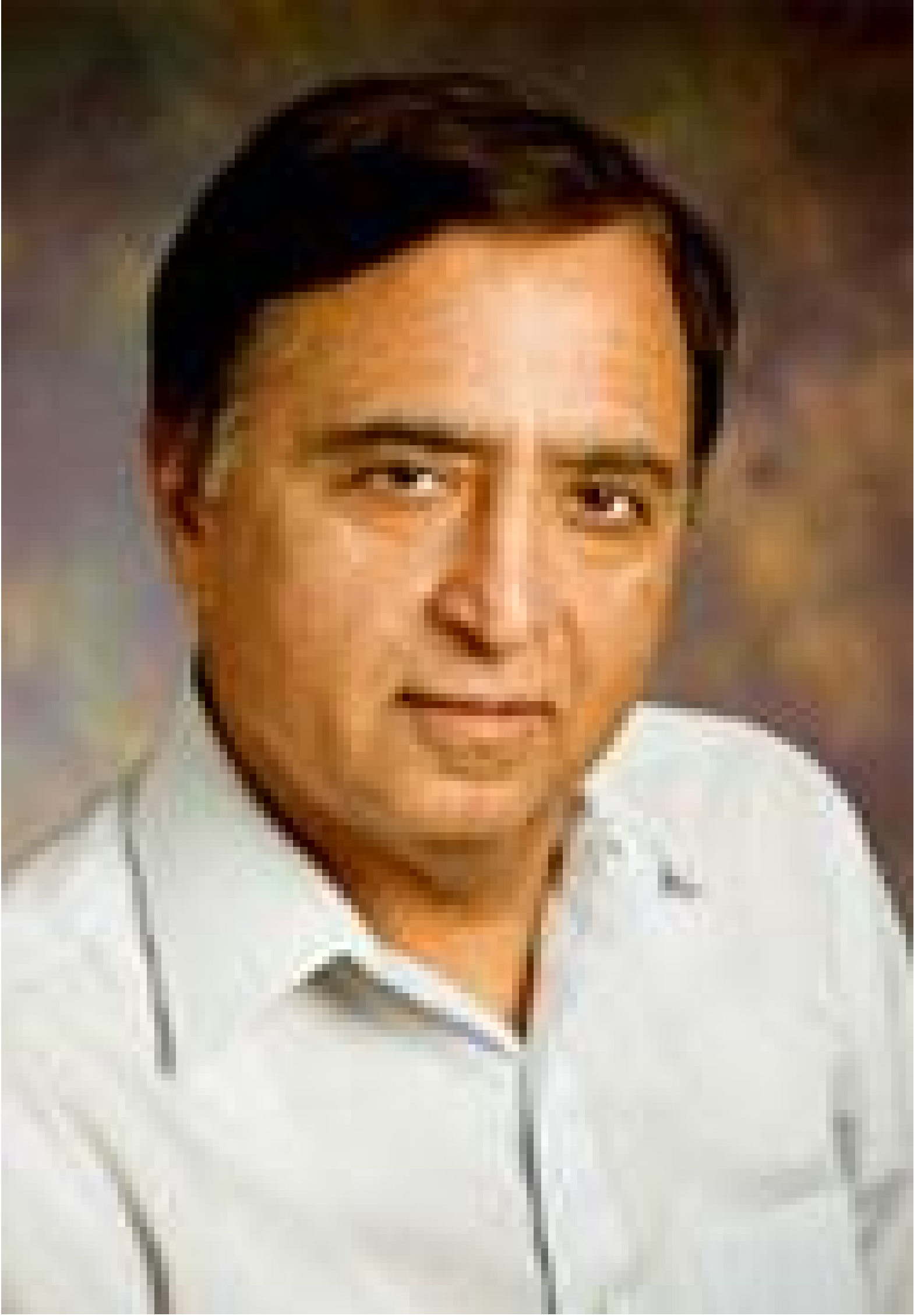}}]{Narendra
Ahuja} received the Ph.D. degree from the University of Maryland, College Park, MD, USA, in 1979. He is the Donald Biggar Willet Professor with the Department of Electrical and Computer Engineering, University of Illinois at Urbana-Champaign, Urbana, IL, USA. Prof. Ahuja is on the Editorial Boards of several journals. He was the Founding Director of the International Institute of Information Technology, Hyderabad, Hyderabad, India, where he continues to serve as a Director International. He is a fellow of the American Association for Artificial Intelligence, the International Association for Pattern Recognition, the Association for Computing Machinery, the American Association for the Advancement of Science, and the International Society for Optical Engineering.
\end{IEEEbiography}

\vspace{-1.2cm}

\begin{IEEEbiography}[{\includegraphics[width=0.9in,clip,keepaspectratio]{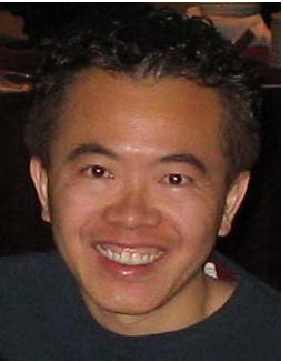}}]{Ming-Hsuan
Yang} is a professor in Electrical Engineering and Computer Science at University of California, Merced. He received the Ph.D. degree in computer science from the University of Illinois at Urbana-Champaign in 2000. 
Yang served as an associate editor of the IEEE Transactions on Pattern Analysis and Machine Intelligence from 2007 to 2011, and is an associate editor of the International Journal of Computer Vision, Image and Vision Computing and Journal of Artificial Intelligence Research. He received the NSF CAREER award in 2012, the Senate Award for Distinguished Early Career Research at UC Merced in 2011, and the Google Faculty Award in 2009. He is a senior member of the IEEE and the ACM.
\end{IEEEbiography}




\end{document}